%% file: main.tex
\documentclass[sigconf,authorversion,nonacm]{acmart}
\usepackage{booktabs} % for professional tables
\usepackage{amsmath}
\usepackage{mathtools}
\usepackage{amsthm}
\usepackage{dsfont}

\usepackage{pgfplotstable}
\usepackage{bbm}
\usepackage{nicefrac}

\usepackage{multirow}
\usepackage{siunitx} % Required for alignment
\usepackage{caption} % For customizing captions

\usepackage{fancyhdr}

\fancypagestyle{firstpage}{
  \fancyhf{} % Clear all header and footer fields
  \fancyfoot[L]{Preprint. Under review.}
}

\usepackage{afterpage}

\usepackage[capitalize,noabbrev]{cleveref}

\theoremstyle{plain}

\theoremstyle{definition}

\theoremstyle{remark}

\usepackage[textsize=tiny]{todonotes}
\AtBeginDocument{%
  \providecommand\BibTeX{{%
    \normalfont B\kern-0.5em{\scshape i\kern-0.25em b}\kern-0.8em\TeX}}}

\setcopyright{acmlicensed}
\copyrightyear{2018}
\acmYear{2018}
\acmDOI{XXXXXXX.XXXXXXX}

\acmConference[Conference acronym 'XX]{Make sure to enter the correct
  conference title from your rights confirmation emai}{June 03--05,
  2018}{Woodstock, NY}
\acmISBN{978-1-4503-XXXX-X/18/06}

\input{notation}
\begin{document}

% \title{An Empirical Evaluation of Calibration in Extreme Multilabel Classification}

\title{Labels in Extremes: How Well Calibrated are Extreme Multi-label Classifiers?}

\author{Nasib Ullah}
\affiliation{%
  \institution{Aalto University}
  \city{Helsinki}
  \country{Finland}}
\email{nasibullah.nasibullah@aalto.fi}

\author{Erik Schultheis}
\affiliation{%
  \institution{Aalto University}
  \city{Helsinki}
  \country{Finland}}
\email{erik.schultheis@aalto.fi}

\author{Jinbin Zhang}
\affiliation{%
  \institution{Aalto University}
  \city{Helsinki}
  \country{Finland}}
\email{jinbin.zhang@aalto.fi}

\author{Rohit Babbar}
\affiliation{%
  \institution{University of Bath}
  \city{Bath}
  \country{UK}}
  \email{rb2608@bath.ac.uk}

\renewcommand{\shortauthors}{Nasib Ullah, et al.}

\begin{abstract}
Extreme multilabel classification (XMLC) problems occur in settings such as related product recommendation, large-scale document tagging, or ad prediction, 
and are characterized by a label space that can span millions of possible labels. There are two implicit tasks that the classifier performs: \emph{Evaluating} each potential label for its expected worth, and then \emph{selecting} the best candidates. For the latter task, only the relative order of scores matters, and this is what is captured by the standard evaluation procedure in the XMLC literature. However, in many practical applications, it is important to have a good estimate of the actual probability of a label being relevant, e.g., to decide whether to pay the fee to be allowed to display the corresponding ad. To judge whether an extreme classifier is indeed suited to this task, one can look, for example, to whether it returns \emph{calibrated} probabilities, which has hitherto not been done in this field. Therefore, this paper aims to establish the current status quo of calibration in XMLC by providing a systematic evaluation, comprising nine models from four different model families across seven benchmark datasets.
As naive application of Expected Calibration Error (ECE) leads to meaningless results in long-tailed XMC datasets, we instead introduce the notion of \emph{calibration@k} (e.g., ECE@k), which focusses on the top-$k$ probability mass, offering a more appropriate measure for evaluating probability calibration in XMLC scenarios.
While we find that different models can exhibit widely varying reliability plots, we also show that post-training calibration via a computationally efficient isotonic regression method enhances model calibration without sacrificing prediction accuracy. Thus, the practitioner can choose the model family based on accuracy considerations, and leave calibration to isotonic regression.
% By studying nine models from four different model families on seven different datasets, we find that instead of the choice of the model family, using post-training calibration via isotonic regression, is a computationally cheap method that improves model calibration without compromising prediction accuracy. 
\end{abstract}

\maketitle

%for preprint
\setlength{\footskip}{25pt}
\thispagestyle{firstpage} % Apply custom page style to the first page

\begin{figure}[ht]
\centering
\includegraphics[width=\linewidth]{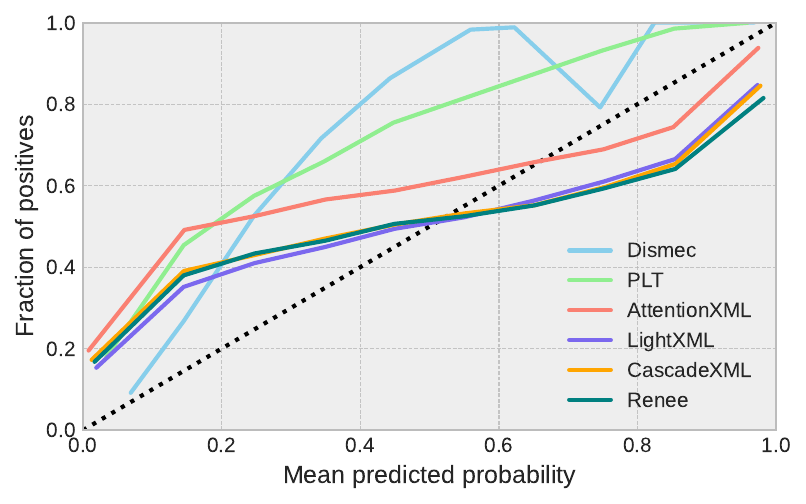} % Adjust the scale as necessary
\caption{Reliability plots at k=3 across different XMLC Models evaluated on the Amazon-670K Dataset. Different models show qualitatively different calibration behaviour.
% While the original models vary considerably in the calibration properties, they are all very well calibrated after isotonic regression.
}
\label{fig:fig1_intro}
\end{figure}

% \begin{figure}[ht]
% \centering
% \includegraphics[width=\linewidth]{figures/ECE_with_P@3_kdd2025.pdf} % Adjust the scale as necessary
% \caption{Comparison of P@3 and ECE@3 across different XMC Models evaluated on the Amazon-670K Dataset, before and after post-training calibration.
% While the original models vary considerably in the calibration properties, they are all very well calibrated after isotonic regression.
% }
% \label{fig:fig1_intro}
% \end{figure}

\section{Introduction}
\label{sec:intro}
\input{parts/intro}

\section{Notation and Background}
\label{sec:background}
\input{parts/background}

\section{Compared Models}
\label{sec:models}
\input{parts/models}

\section{Experiments}
\label{sec:experiments}
\input{parts/experiments}

\section{Results and Insights} 
\label{sec:discussion}
\input{parts/discussion}

\section{Related Work}
\label{sec:related_work}
\input{parts/relatedwork}

\section{Conclusion}
With the growing practical applications of modern deep learning models, studying their calibration properties has been a vibrant research topic recently. 
In the domain of XMLC, while dozens of extreme multi-label classification models have been proposed over the last decade, there does not exist a study to systematically analyse their calibration properties, a gap which we address in this paper.
%In this paper, we studied the calibration properties of these XMLC model families.
%However, in many practical applications, it is actually important to have a good estimate of the actual probability of a label being relevant, e.g., to decide whether to pay the fee to be allowed to display the corresponding ad. To judge whether an extreme classifier is actually suited to this task, one can look, for example, to whether it returns \emph{calibrated} probabilities, which has hitherto not been done in this field. 
In particular, we established the current status quo of calibration in XMLC by providing a systematic evaluation of a wide variety of state-of-the-art XMLC algorithms on common benchmark datasets. 
We find that instead of the choice of the model family, using post-training calibration via isotonic regression, is a computationally cheap method that improves model calibration without compromising prediction accuracy. 
As a future work, we aim to study the calibration properties at the level of individual labels, and investigate if that can be achieved in a computationally tractable manner.

% \begin{acks}
% \end{acks}

\bibliographystyle{ACM-Reference-Format}
\bibliography{sample-base}

\appendix

% \section{Appendix} 
% \label{sec:appendix}
\input{parts/appendix}

\end{document}

%% file: notation.tex
\newcommand{\instancespace}{\mathcal{X}}
\newcommand{\rinstance}{X}
\newcommand{\sinstance}{x}
\newcommand{\rlabelvec}{\boldsymbol{Y}}
\newcommand{\rpredvec}{\hat{\boldsymbol{Y}}}

\newcommand{\numlabels}{m}
\newcommand{\indicator}{\mathds{1}}
\newcommand{\classifier}{\phi}
\newcommand{\scorer}{\psi}
\newcommand{\binsurrogate}{\ell}
\newcommand\expect{\mathds{E}}
\newcommand\prob{\mathds{P}}
\newcommand\ece{\operatorname{ECE}}
\newcommand\acc{\operatorname{acc}}
\newcommand\conf{\operatorname{conf}}
\newcommand\probability[1]{\prob[#1]}
\newcommand\expectation[1]{\expect[#1]}
\newcommand\set[1]{\{#1\}}
\newcommand\given{\mid}

%% file: parts/intro.tex
%!TEX root = ../main.tex

In several domains, such as large scale document classification~\cite{Dekel_Shamir_2010,mencia2008efficient,partalas2015lshtc}
or a wide array of recommendation tasks~\cite{Agrawal_et_al_2013}, the available label
space can be as big as several million labels, a problem referred to as
\emph{extreme
multilabel classification} (XMLC).
%---leading to an \emph{extreme
%multilabel classification} (XMLC) problem.
As a consequence, the number of labels can be in the same order of magnitude as there
are training instances, leading to many labels being tail labels~\cite{adamic2002zipf,babbar2019data}, that is, labels
that have very few positive training instances.

The predominant paradigm for XMLC models currently is to handle the problem
in a  \emph{one-vs-all} (OvA)~\cite{menon2019multilabel, babbar2017dismec, babbar2019data, schultheis2022speeding} fashion: For each
label separately, a score is calculated, and the highest-scoring candidates are
used as the prediction. In practice, several shortlisting   approaches are used so that only scores for a small subset of labels actually need to be evaluated to keep the computational cost tractable\citep{Jasinska-Kobus_et_al_2020,prabhu2018parabel,jiang2021lightxml,kharbanda2022cascadexml}.
The typical evaluation procedure then looks at the
fraction of the $k$ highest scoring labels that are actually
correct (precision-at-$k$, P@k), potentially with different weights given to the 
different positions in the ranking (nDCG@k). A wide range of such evaluations 
is collected in \cite{Bhatia16}.

However, just scoring the most likely labels highest is not a sufficient
criterion for a successful method in many practical settings---instead, one
would like to also get an accurate assessment of the confidence the model has in
its prediction, that is, the model should return \emph{calibrated} probabilities.
This means that the model's subjective probability estimates should coincide with
the actual label frequencies, e.g., in  70\% of the cases where the model
assigns a probability of 70\%, the label should actually be in the ground-truth.
In that sense, it provides a measure for the \emph{reliability} of a given prediction.\looseness=-1

The need for calibrated confidences is obvious in safety-critical applications,
such as when extreme classification methods are used to classify
diseases~\citep{icd-classification,almagro2020icd,duque2021keyphrase,LIU2023102662}.
But even in less high-stakes environments, calibrated probability estimates can
be highly beneficial. In online advertising~\citep{prabhu2018parabel}, an
advertiser might not just want to know which ad is most likely to be successful,
but also what this success probability is, in order to form a maximum value they
would be willing to pay to have this ad displayed. While XMC methods usually retrieve
a top-$k$ ranking, one might also decide to use less than $k$ elements if their
probabilities are too low. Finally, good probability estimates could even be 
helpful to improve XMC learning in itself. \citet{buvaneshenhancing} propose
to use a teacher model in order to perform knowledge distillation, showing that
the reduced training variance can lead to considerable improvements in classifier
performance. If the teacher model is not well calibrated, however, it will send
misleading signals to its student model.

Many XMLC methods use the binary cross-entropy (BCE) loss as their optimizing
criterion, which means that the assigned scores can carry the interpretation of
probability. In particular, as a proper
loss~\citep{savage1971elicitation,buja2005loss,winkler_good_1968}, it is often expected to
directly lead to calibrated probabilities, which is not guaranteed in
general~\citep{blasiok2023when}. More specifically, different model classes
might have different calibration properties, even if they are trained with the
same loss function. For example, it has been observed that early deep networks
yielded well-calibrated classifiers \cite{niculescu2005predicting}, but
subsequent developments, such as ResNet \cite{he2016deep}, produced models that were
overconfident in their predictions \cite{guo2017calibration}. Interestingly, at
least for image classification, the recent shift from convolutional to
transformer-based architectures seems to have restored calibration
\cite{minderer2021revisiting}.

A similar evaluation is currently lacking in the field of XMLC. To the best of
our knowledge, \citet{Jiang2023} is the only paper that thoroughly investigates
the question of uncertainty in XMLC predictions, by adapting ensemble-based
uncertainty quantification methods to the large label-space setting.
In this work, in contrast, we try to follow in the footsteps of \citep{niculescu2005predicting,guo2017calibration,minderer2021revisiting}
and provide a characterization of calibration properties of a wide variety of
XMLC methods.\looseness=-1

To that end, we define the notion of calibration \emph{at-$k$}, that is, we
measure calibration only on the $k$ labels that are ranked highest by the predictor.
This is necessary to prevent the calibration measure from being overwhelmed by
the vast number of easy negative labels that are characteristic of XMLC problems.

We evaluate this measure for shallow linear models~\cite{babbar2019data,babbar2017dismec}, probabilistic
label-trees~\citep{Jasinska-Kobus_et_al_2020}, deep
models~\citep{You_et_al_2019,liu2017deep,kharbanda2023inceptionxml} and
transformer-based
models~\citep{zhang2021fast,jiang2021lightxml,kharbanda2022cascadexml}. This
evaluation also spans several data modalities: tf-idf \cite{jain2016extreme} based
sparse features, bag-of-word representations, long- and short text
sequences, as well as the more recent works taking into account label
features~\citep{NGAME,Dahiya21b,jain2023renee,kharbanda2024learning}.

Overall, we find the results to be quite varied (cf.~\autoref{fig:fig1_intro}): In some situations, the models
are overconfident, sometimes they are underconfident, often depending on the
dataset. Because our calibration measure only needs access to the $k$-highest
scoring predictions, it is computationally feasible to perform post-training
calibration, e.g., using isotonic regression~\citep{isotonic}, which takes only
a few minutes once top-$k$ predictions have been generated. Across methods and
data-sets, we see that such explicit calibration can drastically reduce the
calibration error to almost zero. As this is a monotonic transformation of scores
identical for all labels, traditional performance measures such as precision-at-$k$
are completely unchanged by this correction.

Thus, if we were to compress this paper into a single take-away sentence, it would
be this:
\begin{quote}
    If one cares at all about the accuracy of probability estimates in extreme
    classification problems, using post-training calibration is a computationally
    cheap method that yields accurate probability estimates without compromising
    the classifier's precision!
\end{quote}

%% file: parts/background.tex
%!TEX root = ../main.tex

%\subsection{Notation}
The XMLC problem is concerned with mapping an instance $\rinstance$ to a \emph{set}
of labels, represented by an indicator vector $\rlabelvec \in \{0, 1\}^{\numlabels}$. Our goal is to learn a
predictor $\classifier \in \mathcal{F}: \instancespace \longrightarrow \{0, 1\}^{\numlabels}$ in such
a way that $\rlabelvec$ and $\rpredvec \coloneqq \classifier(\rinstance)$ are as similar as possible. 

\subsection{Losses and Reductions}
There are many ways to compare $\rlabelvec$ and $\rpredvec$, for example subset-0-1 loss
$\indicator(\rlabelvec = \rpredvec)$, Hamming loss, precision, recall, or (instance-wise) F-measure \citep{jasinska_extreme_2016}. 
Often, this function decomposes into a sum over individual labels, as in the case of 
precision or Hamming-loss. In these cases, one can show that a one-vs-all (OvA) reduction
\citep{menon2019multilabel,wydmuch_no-regret_2018} results in a consistent classification procedure: 
Instead of predicting discrete labels using the classifier $\classifier$, predict
the probability for each label independently using a \emph{scorer} $\scorer: \instancespace \longrightarrow R^{\numlabels}$,
and evaluate these scores using a proper \citep{buja2005loss} binary
loss function $\binsurrogate$. Thus, the task becomes to minimize the expected surrogate loss
\begin{equation}
    \min_w \expect \Bigl[\sum_j \binsurrogate(Y_j, \scorer_j(\rinstance; w)) \Bigr] \,.
\end{equation}

Generally, we can decompose the model $\classifier$ into the scorer $\scorer$
and a link function $\sigma: R \rightarrow [0, 1]$, such that we get
probabilistic predictions $\rpredvec \sim \classifier = \sigma \circ \scorer$.
In the case of binary cross-entropy, the scorer generates \emph{logits}, and the
link function is the sigmoid (logistic) function. The introduction of a link
function allows us to turn extreme-classification models that only generated scores, such as
Dismec~\citep{babbar2017dismec}, into models that provide actual probabilities.

In addition to the OvA reduction, instead of decomposing the problem into one binary task for each label, one can also decompose each instance into one multi-class task for each positive label, called a pick-all-labels reduction \citep{wydmuch_no-regret_2018}.
This will still result in consistent classification regarding the popular precision-at-$k$ metric \citep{menon2019multilabel}, which only requires predictions to be ordered correctly within each instance.
In particular, when using the softmax loss, in the infinite data limit, predictions for an instance $X$ will converge to~\citep{wydmuch_no-regret_2018,menon2019multilabel} 
\begin{equation}
    \hat{Y}_j(X) = \prob[Y_j=1 \mid X] \left( \sum_i \prob[Y_i=1 \mid X] \right)^{-1} \,
\end{equation}
i.e., there is an instance-dependent scaling factor that prevents the result from being calibrated.

\subsection{Calibration and its estimators}
\label{sec:calibration}
In a binary setting, a model is calibrated if its reported confidences match
the observed frequencies, 
\begin{equation}
    \prob[Y = 1| \phi(\rinstance) = p] = p \quad \forall p \in [0, 1] \,. \label{eq:binary-calibration}
\end{equation}
Calibration alone is not a sufficient criterion for
a good model. For example, $\phi(\rinstance) = \pi \coloneqq \expect[Y]$, the classifier that just returns
the class prior, is calibrated, but of little use.

The calibration condition \eqref{eq:binary-calibration} is an idealized situation that
is not realized in practice. Therefore, we are interested in measuring how much a given
classifier $\phi$ deviates from this ideal. This can be visualized in calibration curves/reliability diagrams~\citep{murphy_reliability_1977},
and summarized in the \textbf{expected calibration error (ECE) }\cite{naeini2015obtaining} \footnote{similar notions that replace the absolute value with maximum or L2 norm exist. 
For a more thorough discussion, see e.g, \citet{vaicenavicius_evaluating_2019,arrieta2022metrics}.}
\begin{equation}
    \text{ECE} = \expect[|\prob[Y = 1| \phi(\rinstance)] - \phi(\rinstance)|] \,.
\end{equation}

This can be generalized to the multilabel setting in several ways. In accordance
with the OvA decomposition, we could consider the labels independently. We define
the \textbf{marginal calibration} as
\begin{equation}
    \forall j  \colon \quad \expect[Y_j | \phi_j(\rinstance) = \eta_j] = \eta_j \,,
\end{equation}
i.e., each individual binary classification shall be calibrated.
The corresponding ECE would be
\begin{equation}
    \ece = \sum_j  \expect\bigl[\bigl|\prob[Y_j = 1| \phi_j(\rinstance)] - \phi_j(\rinstance)  \bigr|  \bigr] \,.
    \label{eq:full-calibration}
\end{equation}
This indicates that the formulation above might not be suitable for XMLC:
As most labels are tail labels, and thus $\phi_j(\rinstance) \ll 1$ for much of the
probability mass, the expectation above will mainly focus on accurately estimating
the unlikely tail labels. In fact, for Eurlex\citep{mencia2008efficient} with Renee classifier, the resuling ECE would be 0.05, indicating almost prefect calibration.

In practice, however, the probabilistic prediction $\phi$ is turned into a discrete decision by predicting the top-$k$ scoring labels as relevant,
and the others as irrelevant. In that case, we might be more interested in having
accurate confidence in these $k$ labels, and not care so much about the rest.
This leads to \textbf{top-$k$ calibration},
\begin{equation}
    \ece@k = \expect \bigl[ \quad \sum_{\mathclap{j \in \text{top}_k \boldsymbol\phi(\rinstance)}} \; |\prob[Y_j = 1| \phi_j(\rinstance)] - \phi_j(\rinstance)| \bigr] \,, \label{eq:ecek}
\end{equation}
which can be seen as a generalization of \emph{confidence classification}~\citep{guo2017calibration} that
coincides for $k=1$. Taking the example from above, with $k=5$  predictions, the ECE becomes 9.25, showing that the error defined in  \eqref{eq:full-calibration} can be quite misleading in XMLC.

Note that marginal calibration, even if it holds for \emph{each} label individually,
does not imply top-$k$ calibration:
Let $\numlabels=2$, $\instancespace = \set{\sinstance_1, \sinstance_2}$, with $\probability{\sinstance_1} = \probability{\sinstance_2} = \nicefrac{1}{2}$.
Consider a classifier and ground truth values as in \autoref{tab:margina-counterexample}. Note that the
presented values do fulfill the item-level calibration condition.
Now, consider the resulting top-1 predictor $\classifier(\sinstance) \coloneqq \text{top}_1(\scorer(\sinstance))$. In this case, we
have $\classifier(\sinstance_1) = 2$, and $\classifier(\sinstance_2) = 1$. Correspondingly,
the classifier only predicts values 0.9 and 0.7. We thus need to check
\begin{align*}
    \expectation{\prob[Y_{\classifier(\rinstance)}] \given \scorer(\rinstance, \classifier(\rinstance)) = 0.9} &= 
    \expectation{\prob[Y_2] \given \rinstance = \sinstance_1} = 0.9 \\ 
    \expectation{\prob[Y_{\classifier(\rinstance)}] \given \scorer(\rinstance, \classifier(\rinstance)) = 0.7} &= 
    \expectation{\prob[Y_1] \given \rinstance = \sinstance_2} = 0.6 \neq 0.7 \,.
\end{align*}
We can see that top-k selection introduces a bias, so that even if the scorer is calibrated
for each label individually, applying it to the highest scoring label breaks calibration.

\begin{table}[t]
    \caption{A counterexample in which marginal calibration does not lead to top-1 calibration.}
    \centering
    \begin{tabular}{ccccc}
    \toprule
    $\sinstance$ & \multicolumn{2}{c}{label 1} & \multicolumn{2}{c}{label 2}\\ 
    & $\prob[Y_1 = 1]$ & $\scorer(\sinstance, 1)$ & $\prob[Y_2 = 1]$ & $\scorer(\sinstance, 2)$ \\  \midrule
    $\sinstance_1$ & 0.8 & 0.7 & 0.9 & 0.9\\
    $\sinstance_2$ & 0.6 & 0.7 & 0.4 & 0.4\\ \bottomrule
\end{tabular}
    \label{tab:margina-counterexample}
\end{table}

For discussions on different notions of calibration for multiclass
problems, we refer the reader to \citet{gupta_top-label_2021,widmann_calibration_2019,vaicenavicius_evaluating_2019}.
% https://arxiv.org/pdf/2109.03480.pdf  Def 2 and 3, maybe;

In order to estimate \eqref{eq:ecek}, a straightforward and commonplace way is
to discretize the predictions into a small number of bins $B_1, \ldots, B_m$ and approximate
\begin{equation}
    \ece \approx \sum_{i=1}^m \frac{|B_i|}{n} \bigl| \acc(B_i) - \conf(B_i)  \bigr| \,,
\end{equation}
where $n = \sum_{i=1}^m |B_i|$ is the total number of test instances, and
$\acc$ denotes the accuracy of the bin $\acc(B_i) = |B_i|^{-1}\sum_{j \in B_i} Y_j$,
i.e., the empirical frequency of positive labels among all classifications with
confidence in a certain range. Here $\conf(B_i) = |B_i|^{-1}\sum_{j \in B_i} \phi(x_j)$, in turn,
is the average confidence of all samples within the bin.

Binning predictions using fixed size bin can be problematic \cite{roelofs2022mitigating}. As a case in point, if the bins
are wide enough to average predictions where the network is overconfident, along with
predictions where it is underconfident, it might report deceptively favourable
results. As one way to counteract these problems, \citet{NixonDZJT19} propose
to switch from fixed-interval binning to quantile binning, that is, choose bin
boundaries such that each bin contains an equal number of samples, resulting in
the \emph{Adaptive Calibration Error (ACE)}.

\subsection{Post-training calibration}
Instead of trying to achieve classifier training that directly results in
calibrated probabilities, one can also try to adjust the generated scores
in a post-training step to achieve calibration, especially when scores are
not limited to the range $[0, 1]$.

\emph{Isotonic regression}~\citep{isotonic} is a non-parametric method that
produces a monotonic function for mapping arbitrary scores into the unit interval, 
by solving the optimization
\begin{equation}
    \begin{aligned}
    & \underset{\theta_1,\ldots,\theta_M}{\text{minimize}}
    & & \frac{1}{M} \sum_{m=1}^{M} \sum_{i=1}^{n} \mathds{1}[a_m \leq \phi(x_i) < a_{m+1}] (\theta_m - y_i)^2 \\
    & \text{subject to}
    & & 0 = a_1 \leq a_2 \leq \ldots \leq a_{M+1} = 1, \\
    &&& \theta_1 \leq \theta_2 \leq \ldots \leq \theta_M \,,
    \end{aligned}
\end{equation}
where \( M \) is the number of intervals, \( a_1, \ldots, a_{M+1} \) are the interval boundaries, and \( \theta_1, \ldots, \theta_M \) are the function values.

\noindent Because of the highly flexible nature of the transformation learned by isotonic 
regression, it can be very effective if enough calibration data is available,
but lead to overfitting if the sample is too small~\citep{niculescu2005predicting}.
As an alternative, \emph{Platt scaling}~\citep{platt1999probabilistic}
parameterizes the transformation using a logistic function $s \mapsto (1 + \exp(a s  + b))^{-1}$,
which has only two tunable parameters $a$ and $b$, and thus is more adequate in scarce data scenarios.

%% file: parts/models.tex
%!TEX root = ../main.tex

\noindent\textbf{Linear Models.}
Arguably, the simplest family of models used in XMLC are linear models \cite{babbar2017dismec,babbar2019data}, which learn a sparse linear classifier
on top of \emph{tf-idf} features. Even though both feature and label space are
gigantic, the sparsity in both the feature and weight vectors allows for efficient
computations and reasonable memory requirements, and the OvA decomposition together
with the \emph{linear} score calculation allow for trivial parallelization, making
these methods highly scalable. Using clever initialization strategies \cite{fang2019fast,schultheis2022speeding},
the computational cost can be further improved.

In their standard formulations, these models use the squared hinge-loss as their
objective, which is a proper loss, but does not directly result in interpretable
probabilities. Therefore, following the way \citet{niculescu2005predicting} handled
SVM classifiers, we squash the predicted scores into the range $[0, 1]$ by a linear
transformation $s' = (s - \min) / (\max - \min)$.

\noindent\textbf{Label-Tree based Models.}
While the wall-clock time requirements for linear models can be brought down by
massive parallelism, but the overall compute cost remains very high. In
label-tree based models, this is alleviated by imposing a tree-structure on the
labels. This allows to skip entire sub-trees which contain only negative labels
during calculations, thus reducing the asymptotic cost from linear in the number
of labels to logarithmic.

In \emph{Probabilistic Label Trees} (PLT), the probability of a label is factorized along the path $n_0, \ldots, n_l$ from tree root to tree leaf (label), in the sense of \cite{Jasinska-Kobus_et_al_2020,jasinska_extreme_2016}
\begin{equation}
    \mathds{P}[n_l \mid X] = \prod_{s=1}^l \mathds{P}[n_s \mid n_{s-1}, X] \,.
\end{equation}
By only performing calculations in subtrees that contain high-probability nodes,
the computational complexity of label-tree based approaches can be made sublinear
in the number of labels.

Additional works in this vein include \emph{Bonsai}~\citep{khandagale2020bonsai}, 
\emph{Parabel}~\citep{prabhu2018parabel}, and \emph{extremeText}~\citep{wydmuch_no-regret_2018}. 
For this paper, we used \emph{NapkinXC}'s PLT implementation\footnote{\url{https://pypi.org/project/napkinxc/}}
with an ensemble of $3$ trees and a maximum leaf count $m=200$ as a representative for this line of work.

\noindent \textbf{Early Deep Models.} Early deep learning approaches for XMLC replace tf-idf and BOW features with word embeddings derived from raw text, utilizing convolutional and recurrent neural networks. XML-CNN \cite{liu2017deep}, the first method in this group, employs one-dimensional convolutional layers to process word embeddings, followed by a low-dimensional hidden layer, and evaluates all labels using binary cross-entropy loss. AttentionXML \cite{You_et_al_2019}, another significant contribution in this category, incorporates an LSTM network with attention mechanism and integrates it with a probabilistic label tree. In this structure, each tree level is trained using a separate LSTM network,  effectively managing the extreme label space. For our study, we use the original implementation\footnote{\url{https://github.com/yourh/AttentionXML}} of AttentionXML \cite{You_et_al_2019}, as it represents one of the strongest baselines in this category.
\looseness=1

% Among the initial, pivotal techniques in this group, XML-CNN \cite{liu2017deep} and AttentionXML \cite{You_et_al_2019} stand out for their innovative contributions. XML-CNN leverages one-dimensional convolutional layers to process word embeddings, culminating in a dense layer tasked with evaluating all labels through binary cross-entropy loss. In contrast, AttentionXML, achieving greater success, implements a probabilistic label tree (PLT) strategy, underpinned by  LSTM network that incorporates attention mechanisms at each hierarchical level, facilitating improved label predictions.
% \todo[inline]{footnote about XML-CNN}
\noindent \textbf{Transformer Models.}
Recent advancements in XMLC have leveraged pre-trained transformer architectures to enhance performance. LightXML \cite{jiang2021lightxml} employs meta-classifiers from a constructed label tree to shortlist labels, effectively implementing dynamic negative sampling for extreme ranking stages. This approach has been further developed in models such as X-Transformer \cite{Chang20}, XR-Transformer \cite{zhang2021fast}, CascadeXML \cite{kharbanda2022cascadexml}, and ELIAS \cite{gupta2022elias}. The core concept in these models revolves around fine-tuning transformer encoders using single or multi-resolution meta-classifiers constructed from probabilistic label trees to manage the extreme label space.\looseness=1

Meta-classifier outputs significantly impact final performance, leading to diverse construction strategies in the literature. For instance, (i) LightXML \cite{jiang2021lightxml} utilizes single-resolution meta-classifiers, whereas CascadeXML \cite{kharbanda2022cascadexml} and XR-Transformer \cite{zhang2021fast} implement multi-resolution approaches. (ii) LightXML \cite{jiang2021lightxml} and CascadeXML \cite{kharbanda2022cascadexml} use predefined features to cluster the label space and construct fixed meta-classifiers throughout training, while ELIAS \cite{gupta2022elias} learns label-to-cluster assignments concurrently with the training objective. While these strategies have improved precision, \emph{their impact on model calibration has not been previously examined.}

In our study, we employed LightXML \cite{jiang2021lightxml} and CascadeXML \cite{kharbanda2022cascadexml}, alongside Renee \cite{jain2023renee}, a recent method enabling efficient end-to-end training of XC models at extreme scale without meta-classifiers.

% LightXML \cite{jiang2021lightxml} leverages a pre-trained transformer architecture for the dual tasks of label recall (via a label cluster tree) and ranking, embodying a novel, end-to-end training methodology. This methodology synergizes recall and ranking processes through a generative cooperative framework, facilitating dynamic negative sampling.
% %\todo[inline]{write about CascadeXML} 
% Another recent transformer-based baseline that we employ in our experimental analysis is CascadeXML \cite{kharbanda2022cascadexml}, which is an end-to-end multi-resolution learning pipeline, capable of harnessing the multi-layered transformer model for attending to different label resolutions attained through a probabilistic label tree.

\noindent \textbf{Label Features based Models.}
The current frontier of XMLC research focuses on incorporating label features—including text, images, and label correlation graphs—to enhance performance metrics such as precision. Two primary approaches have emerged in this domain: (i) Dual encoder-based siamese architectures, where data points and labels are trained via contrastive loss followed by OvA classifiers. Notable methods in this category include SiameseXML \cite{Dahiya21b}, ECLARE \cite{Mittal21b}, GalaXC \cite{GalaXC}, DeepXML \cite{Dahiya21}, MUFIN \cite{Mittal22}, and NGAME \cite{NGAME}. Our study focuses on NGAME \cite{NGAME}, which employs negative mining-aware mini-batching, and GalaXC \cite{GalaXC}, which utilizes graph convolutional networks. (ii) Label feature augmentation, where label features are treated as additional data points with either hard or soft labels. Renee \cite{jain2023renee}  employs hard labels for augmented label features, while Gandalf \cite{kharbanda2024learning}  uses soft labels.\looseness=1

Our primary focus is to \emph{investigate how label feature-based models, which boost performance, affect calibration}. Additionally, it is important to \emph{study how different approaches to incorporating label features influence calibration properties}. For instance, two-tower based methods typically use contrastive loss, which is not suitable for calibration. Both Renee \cite{jain2023renee} and Gandalf \cite{kharbanda2024learning} use symmetric binary cross-entropy loss, but Gandalf employs soft labels, acting as label smoothing, a technique known to be beneficial for calibration \cite{muller2019does}.\looseness=-1

% We consider two state-of-the-art models : (i)
% Ngame \citep{NGAME}, which  employs a two-tower pre-training stage by applying contrastive learning between an input text and its corresponding label features, and (ii) \textsc{GalaXC} \citep{GalaXC} is motivated by graph convolutional networks wherein it creates a combined query-label bipartite graph to aggregate predicted instance's neighbourhood.

%% file: parts/experiments.tex
%!TEX root = ../main.tex

\begin{table*}
\centering
\caption{Calibration error metrics (at $k=5$, detailed in 
\autoref{tab:main_table_extended}) on four datasets for a range of XMLC algorithms. Two consecutive rows, such as DiSMEC and DiSMEC-I, represent the measures on the vanilla version of the algorithm and that obtained by re-calibration via Isotonic regression respectively.}
\label{tab:main_table}
\begin{tabular}{
  @{}
  l
  l
  S[table-format=2.2]
  S[table-format=2.2]
  S[table-format=2.2]
  S[table-format=2.2]
  S[table-format=2.2]
  S[table-format=2.2]
  S[table-format=2.2]
  S[table-format=2.2]
  S[table-format=2.2]
  S[table-format=2.2]
  S[table-format=2.2]
  S[table-format=2.2]
  @{}
}
\toprule
 & \textbf{Calibration Measure} &  \multicolumn{1}{|c}{\textbf{ECE@k}} & {\textbf{Brier}} & {\textbf{ACE}} & \multicolumn{1}{|c}{\textbf{ECE@k}} & {\textbf{Brier}} & {\textbf{ACE}} & \multicolumn{1}{|c}{\textbf{ECE@k}} & {\textbf{Brier}} & {\textbf{ACE}} & \multicolumn{1}{|c}{\textbf{ECE@k}}& {\textbf{Brier}} & {\textbf{ACE}} \\
\midrule
 & \multicolumn{1}{c}{\textbf{Dataset}} & \multicolumn{3}{|c}{\textbf{Eurlex-4K}} & \multicolumn{3}{|c}{\textbf{AmazonCat-13K}} & \multicolumn{3}{|c}{\textbf{Wiki-500K}} & \multicolumn{3}{|c}{\textbf{Amazon-670K}} \\ 
\midrule
  & \multicolumn{1}{c}{{DiSMEC}} & \multicolumn{1}{|c}{32.86} & 30.79 & 32.86  & \multicolumn{1}{|c}{58.38} & 55.32 & 58.38 & \multicolumn{1}{|c}{-} & \multicolumn{1}{c}{-}  & \multicolumn{1}{c}{-}  & \multicolumn{1}{|c}{17.76} & 20.98 & 17.76 \\
& \multicolumn{1}{c}{{DiSMEC-I}}  & \multicolumn{1}{|c}{0.05} & 18.39 & 0.36  & \multicolumn{1}{|c}{0.05} & 1.24 & 0.03 & \multicolumn{1}{|c}{-} & \multicolumn{1}{c}{-} & \multicolumn{1}{c}{-} & \multicolumn{1}{|c}{0.04} & 1.57 & 0.03 \\
\midrule
& \multicolumn{1}{c}{{PLT}}  & \multicolumn{1}{|c}{9.67} & 2.03 & 9.71  & \multicolumn{1}{|c}{4.63} & 13.05 & 4.65 & \multicolumn{1}{|c}{13.3} & 17.59 & 13.3 & \multicolumn{1}{|c}{19.7} & 20.69 & 19.7 \\
& \multicolumn{1}{c}{{PLT-I}}  & \multicolumn{1}{|c}{0.96} & 19.22& 0.51  & \multicolumn{1}{|c}{0.05} & 12.68 & 0.02 & \multicolumn{1}{|c}{0.04} & 15.45 & 0.02 & \multicolumn{1}{|c}{0.04} & 15.75 & 0.04 \\
\midrule
& \multicolumn{1}{c}{{AttentionXML}}  & \multicolumn{1}{|c}{10.9} & 19.51 & 10.63 & \multicolumn{1}{|c}{2.65} & 11.36 & 2.59 & \multicolumn{1}{|c}{10.01} & 16.01 & 10.01 & \multicolumn{1}{|c}{16.11} & 19.14 & 15.71 \\
& \multicolumn{1}{c}{{AttentionXML-I}}  & \multicolumn{1}{|c}{0.32} &18.12 & 0.78  & \multicolumn{1}{|c}{0.04} & 11.24 & 0.02 & \multicolumn{1}{|c}{0.05} & 14.56 & 0.03 & \multicolumn{1}{|c}{0.02} & 15.04 & 0.26\\
\midrule
& \multicolumn{1}{c}{{LightXML}}  & \multicolumn{1}{|c}{13.28}& 19.65 & 13.44 & \multicolumn{1}{|c}{4.16} & 10.83 & 4.07 & \multicolumn{1}{|c}{10.23} & 15.95 & 10.15 & \multicolumn{1}{|c}{12.86} & 17.48 & 12.89 \\
& \multicolumn{1}{c}{{LightXML-I}}  & \multicolumn{1}{|c}{0.26} & 17.29 &0.74  & \multicolumn{1}{|c}{0.01} & 10.52 & 0.02 & \multicolumn{1}{|c}{0.03} & 14.43 & 0.05 & \multicolumn{1}{|c}{0.05} & 15.35 & 0.04 \\
\midrule
& \multicolumn{1}{c}{{CascadeXML}}  & \multicolumn{1}{|c}{12.76} & 19.75 & 12.5  & \multicolumn{1}{|c}{8.28} & 11.43& 9.79 & \multicolumn{1}{|c}{12.26} & 16.88 & 12.17 & \multicolumn{1}{|c}{14.54} & 18.18 & 14.41 \\
& \multicolumn{1}{c}{{CascadeXML-I}}  & \multicolumn{1}{|c}{0.43} & 17.85 & 0.47 & \multicolumn{1}{|c}{0.13} & 10.31 & 0.03  & \multicolumn{1}{|c}{0.03} & 14.69 & 0.04 & \multicolumn{1}{|c}{0.04} & 15.4 & 0.04 \\
\midrule
& \multicolumn{1}{c}{{Renee}}  & \multicolumn{1}{|c}{9.25}& 18.99 & 9.18 & \multicolumn{1}{|c}{3.6} & 11.05 & 3.52  & \multicolumn{1}{|c}{10.13} & 16.15 & 10.04 & \multicolumn{1}{|c}{14.94} & 18.64 & 14.78 \\
& \multicolumn{1}{c}{{Renee-I}}  & \multicolumn{1}{|c}{0.59}& 17.78 & 0.65  & \multicolumn{1}{|c}{0.06} & 10.87 & 0.02 & \multicolumn{1}{|c}{0.04} & 14.91 & 0.04 & \multicolumn{1}{|c}{0.05} & 15.81 & 0.03 \\
\bottomrule
\end{tabular}
\end{table*}

\noindent \textbf{Datasets.} Our investigation utilizes an extensive array of benchmark datasets available at the Extreme Classification Repository \cite{Bhatia16}, which features a rich variety of data representations, including tf-idf vectors, bag-of-words models, and both long and short text sequences, complemented by label metadata presented as label text. The utility of these datasets spans several domains, from product recommendation systems, exemplified by Amazon-670K, LF-Amazon-131K \cite{mcauley2013hidden}, LF-AmazonTitles-131K \cite{mcauley2013hidden}, AmazonCat-13K\citep{mcauley2013hidden}, to the inference of related articles, as in LF-WikiSeeAlso-320K, and to classification tasks, such as those presented by Wiki-500K and EurLex-4K\citep{mencia2008efficient}. The datasets' statistical profiles are delineated in Table \ref{tab:dataset_statistics}.

\begin{table}[bt]
\centering
\captionsetup{width=\columnwidth} % Adjusts caption to be within half column width
\caption{Statistics of Datasets with and without Label Features. This table presents a comparison across various datasets, detailing the total number of training instances (N), the unique labels (L), the number of test instances ($N^{'}$), the average label count per instance ($\overline{L}$), and average data points per label ($\hat{L}$).}
\resizebox{\columnwidth}{!}{ % Resizes the table to fit half column width
\begin{tabular}{
  @{} % Removes extra column spacing at the edges
  l
  S[table-format=7.0]
  S[table-format=6.0]
  S[table-format=6.0]
  S[table-format=2.2]
  S[table-format=2.2]
  @{} % Removes extra column spacing at the edges
}
\toprule
{Dataset} & {N} & {L} & {\(N^{'}\)} & {\(\overline{L}\)} & {\(\hat{L}\)} \\
\midrule
\multicolumn{6}{c}{Datasets without Label Features} \\
\midrule
Eurlex-4K & 15449 & 3956 & 3865 & 5.30 & 20.79 \\
AmazonCat-13K & 1186239 & 13330 & 306782 & 5.04 & 448.57 \\
Wiki-500K & 1779881 & 501070 & 769421 & 4.75 & 16.86 \\
Amazon-670K & 490449 & 670091 & 153025 & 5.45 & 3.99 \\
Amazon-3M & 2812281 & 1717899 & 742507 & 36.17 & 31.64 \\
\addlinespace
\midrule
\multicolumn{6}{c}{Datasets with Label Features} \\
\midrule
LF-Amazon-131K & 294805 & 131073 & 134835 & 5.15 & 2.29 \\
LF-WikiSeeAlso-320K & 693082 & 312330 & 177515 & 4.68 & 2.11 \\
LF-AmazonTitles-131K & 294805 & 131073 & 134835 & 5.15 & 2.29 \\
\bottomrule
\end{tabular}
}
\label{tab:dataset_statistics}
\end{table}

\begin{figure*}
\centering
\includegraphics[width=\linewidth]{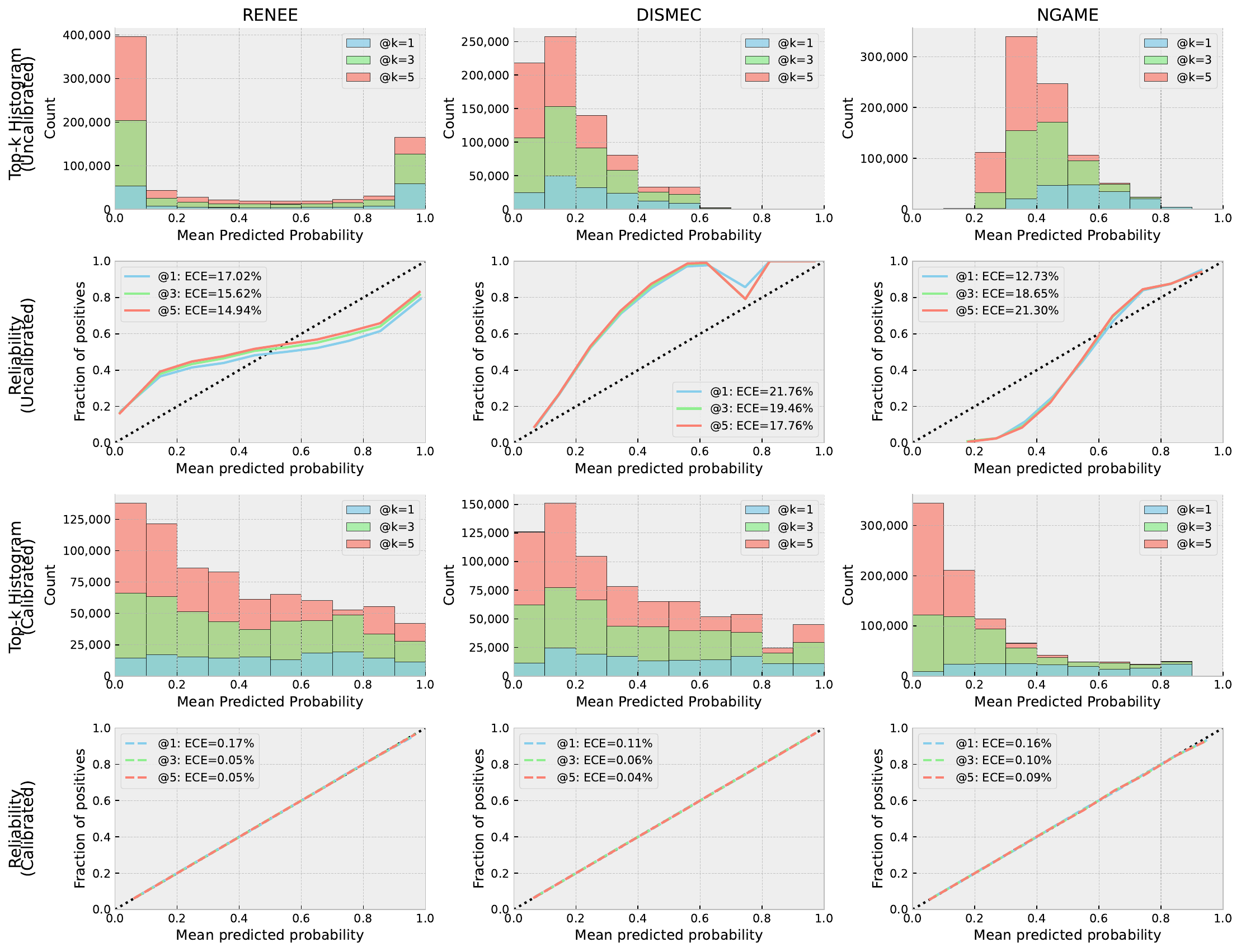} % Adjust the scale as necessary
\caption{Calibration Effects on Top-k Prediction Probabilities and Reliability. Top-$k$ prediction histograms (top row: uncalibrated, third row: calibrated) and reliability plots (second row: uncalibrated, bottom row: calibrated) for RENEE \cite{jain2023renee} (representative of Transformer and deep XMLC models), DISMEC \cite{babbar2017dismec} (representative of linear and PLT-based models) on Amazon-670K, and NGAME \cite{NGAME} (representative of two-tower label feature-based XMLC models) on LF-WikiSeeAlso-320K. 
%The bottom row (\textit{notice the difference in the scale of the y-axis}) depicts various measures pre- and post- calibration. Not only that calibration error decreases substantially but the prediction accuracy (P@5) remains unchanged which affirms the efficacy of re-calibration
}
%\todo{Instead of ECE and Brier, can we have before/after isotonic regresion here?}
\label{fig:fig1_experiments_main_calib_curve}
\end{figure*}

\noindent\textbf{Metrics for Measuring Calibration} We evaluate $\ece@k$ for values $k \in \{1, 3, 5\}$, to align with commonly used evaluation benchmarks in XMLC. We use a fixed bin size of 10 for this calculation. For implementing ACE \cite{NixonDZJT19}, we followed the official repository of \citet{pmlr-v216-matsubara23a}.

\begin{figure}[ht]
\centering
\includegraphics[width=\linewidth]{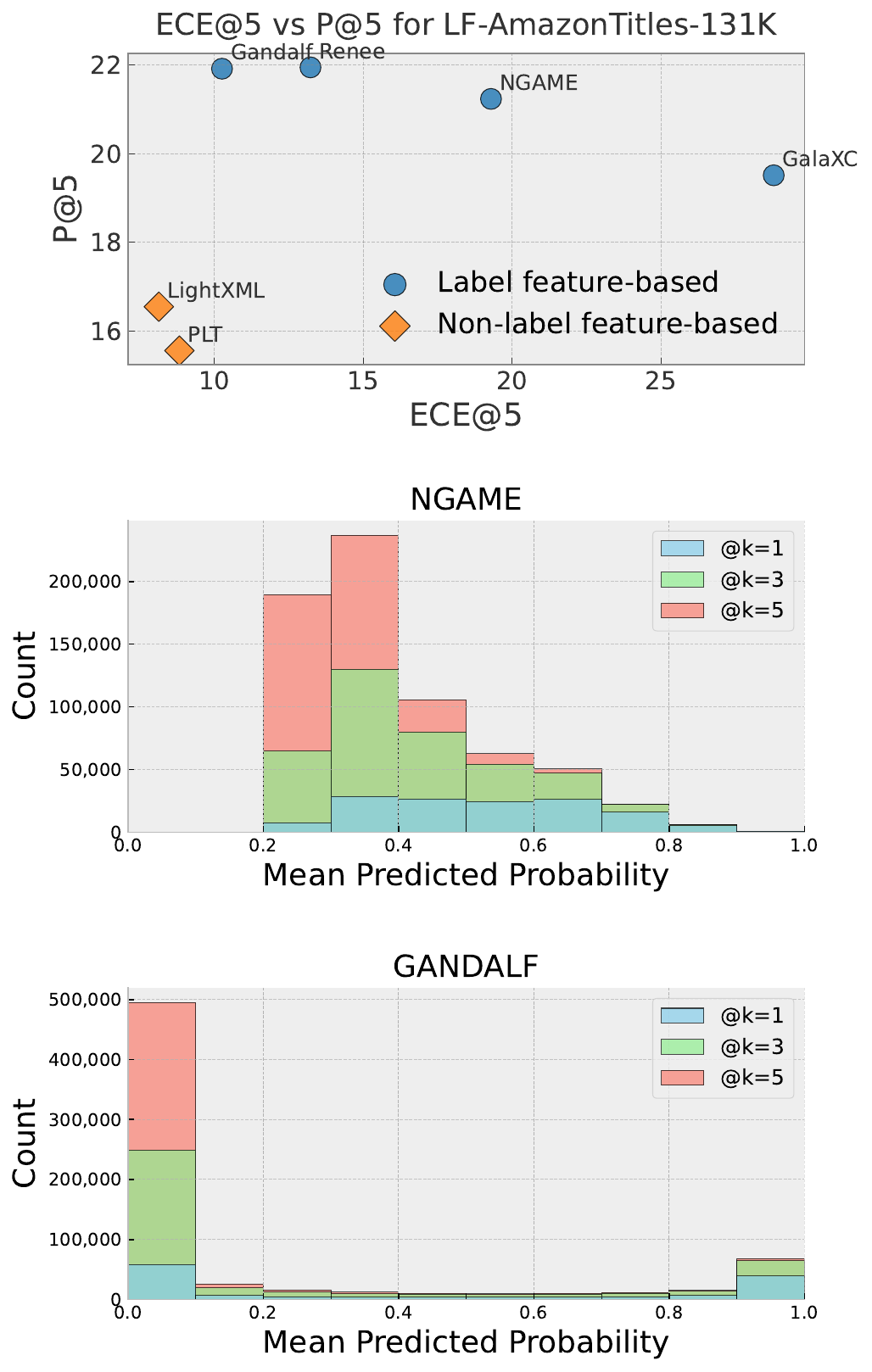} % Adjust the scale as necessary [label_feature_effect.pd]
\caption{First Row: Comparative analysis of calibration error metrics and P@5 performance between methods with label features (GalaXC \cite{GalaXC}, NGAME \cite{NGAME} , Renee \cite{jain2023renee}, Gandalf \cite{kharbanda2024learning}) and without label features (PLT, LIGHTXML \cite{jiang2021lightxml}) on the LF-AmazonTitles-131K dataset. Second and Third Rows: Top-$k$ probability prediction patterns for NGAME \cite{NGAME} and Gandalf \cite{kharbanda2024learning} }
\label{fig:label_vs_nonlabel}
\end{figure}

 \noindent\textbf{Brier Score} The Brier Score~\citep{Brier1950} serves as complementary metrics to Expected Calibration Error (ECE). Unlike ECE/ECE@$k$, Brier Score does not directly measure calibration; instead, it provides a broader assessment of probabilistic forecasts' quality. The Brier Score quantifies the mean squared deviation between predicted probabilities and the observed outcomes,
\begin{equation}
  \operatorname{BS} = n^{-1} \sum (\phi(x_i) - y_i)^2\,.
\end{equation}
It can be decomposed into three distinct components: Uncertainty, Reliability, and Resolution \cite{1973JApMe12595M}.
Consequently, a lower Brier Score does not unambiguously indicate a well-calibrated model; it may also reflect a model with superior discriminative ability that utilizes a more extensive feature set, potentially overshadowing calibration issues. For computing the Brier score, we use scikit-learn\footnote{\href{https://scikit-learn.org/stable/modules/generated/sklearn.metrics.brier_score_loss.html}{scikit-learn.org/stable/modules/generated/sklearn.metrics.brier\_score\_loss.html}}.

%\noindent\textbf{Under and Overconfidence with EUCE@k and EOCE@k}  For domain-specific applications, distinguishing between under and over calibration errors is more informative than a composite ECE. Consider medical diagnostics~\citep{icd-classification,almagro2020icd,duque2021keyphrase,LIU2023102662}, where under-calibration at lower probability thresholds could have critical consequences, thereby necessitating a more focused assessment of underestimation errors. To address this, we introduce the Expected Underconfidence Error (EUCE@k) and Expected Overconfidence Error (EOCE@k), which sum up the negative and positive contributions in the
%ECE, respectively:
%\[ \operatorname{EUCE}\!@k = \sum_{i=1}^{m} \frac{\left|B_{i} \right|}{n} \max((\text{acc}(B_{i})-\text{conf}(B_{i}),0)  \]
%\[ \operatorname{EOCE}\!@k = \sum_{i=1}^{m} \frac{\left|B_{i} \right|}{n} \max((\text{conf}(B_{i})-\text{acc}(B_{i}),0)  \]
%In our analysis, \( B_i \) denotes the \( i \)-th bin, among a total of \( m \) bins. Here, \( n \) signifies the aggregate number of data points.

\noindent \textbf{Calibration Methods} 
In the recalibration process applied to the test dataset, we employed k-fold cross-validation. The test set was divided into k disjoint subsets. For each fold, an isotonic regression model was trained on the union of k-1 subsets and then used to predict recalibration probabilities for the excluded subset. This procedure was iterated such that each fold served as the validation set once, ensuring that recalibration was assessed on all portions of the test data.

%Label vs Non Label feature based methods. Label Scaling 

%\todo[inline]{Label Features: The Key to Better Calibration?}
%\todo[inline]{One figure showing calibration curves for one dataset}
%\todo[inline]{Table with ECE/ECE-at-k}
%\todo[inline]{Post-training calibration results}
%\todo[inline]{Evaluate ECE separately for head and tail}

%% file: parts/discussion.tex
%!TEX root = ../main.tex

\noindent \textbf{Observations} In our empirical evaluation of various XMLC algorithms across multiple datasets, we observed distinct calibration characteristics. Tables \ref{tab:main_table} and \ref{tab:label_feature_table} summarize the calibration error metrics—ECE@$K$, ACE, and Brier scores—for label feature-based and non-label feature-based approaches, respectively. Figure \ref{fig:fig1_experiments_main_calib_curve} provides complementary visual assessments of top-k predicted probabilities and reliability diagrams for both uncalibrated and calibrated versions.
Our analysis revealed three primary patterns:
\begin{enumerate}
    \item \emph{Transformer-based and deep models} exhibit predictions skewed towards extremes, particularly for larger datasets. Reliability plots indicate overestimation at high probabilities and underestimation at lower ranges (Figure \ref{fig:fig1_experiments_main_calib_curve}, left column), resulting in a classical overconfidence curve.
    \item \emph{Linear models and PLT-based methods} demonstrate uncalibrated top-$k$ predictions biased towards lower probabilities (Figure \ref{fig:fig1_experiments_main_calib_curve}, middle column) and consistent under-confidence in their reliability diagrams.
    \item \emph{Label feature-based methods} employing two-tower architecture show uncalibrated predicted probabilities concentrated in the mid-range, leading to a reliability plot with opposite shape compared to the overconfident transformers.
\end{enumerate}

Despite these variations, post-hoc calibration via isotonic regression yields more uniform characteristics across all model types. Calibrated models exhibit a more even distribution of predicted probabilities (\autoref{fig:fig1_experiments_main_calib_curve}, third row), improved alignment in reliability diagrams (\autoref{fig:fig1_experiments_main_calib_curve}, fourth row), and enhanced quantitative measures of calibration (\autoref{tab:main_table} and \ref{tab:label_feature_table}).

% \todo[inline]{Provide an interpretation for the results in the previous section}
% \todo[inline]{As we are not proposing a new method, this section should be a lot larger than in "typical" papers}
% \todo[inline]{Provide an outlook on new research problems that open up when looking at calibration in XMLC: calibrating individual labels, calibration under missing labels}

\noindent \textbf{Impact of Label Features on calibration:}  In recent XMLC practice, incorporating label features has proven instrumental in enhancing prediction performance compared to approaches that ignore this information. To assess their impact on calibration, \autoref{fig:label_vs_nonlabel} presents a precision vs. ECE@k plot comparing methods that utilize label features against those that do not. Interestingly, methods such as LightXML \cite{jiang2021lightxml} and PLT, which do not exploit label features, exhibit better calibration than label feature-based approaches like NGAME \cite{NGAME}, GalaXC \cite{GalaXC}, and Renee \cite{jain2023renee}.

\noindent Among label feature-based methods, we observe varying degrees of calibration, suggesting that the strategy for incorporating label features significantly influences calibration quality. \autoref{tab:label_feature_table} reveals that methods employing label features as augmented data points with symmetric binary cross-entropy loss (e.g., Renee \cite{jain2023renee}, Gandalf \cite{kharbanda2024learning}) achieve superior calibration compared to two-tower approaches using contrastive loss. Notably, Gandalf's soft label augmentation outperforms Renee's hard label augmentation in terms of calibration. \autoref{fig:label_vs_nonlabel} further illustrates that augmentation-based label feature methods exhibit top-k prediction patterns similar to transformer and deep model approaches.

% Label Feature Methods Table

\begin{table}
%\centering
\caption{Calibration error metrics (at $k=5$, detailed in \autoref{tab:LF_table_extended}) for label features based XMLC algorithms. Two consecutive rows, such as NGAME and NGAME-I, represent the measures on the vanilla version of the algorithm and that obtained by post-hoc calibration via Isotonic regression. }
\label{tab:label_feature_table}
\begin{tabular}{
  @{}
  l
  l
  S[table-format=2.2]
  S[table-format=2.2]
  S[table-format=2.2]
  S[table-format=2.2]
  @{}
}
\toprule
  \multicolumn{2}{l}{\textbf{Calibration }} &  \multicolumn{1}{|c}{\textbf{ECE@k}} & {\textbf{Brier}}  & \multicolumn{1}{|c}{\textbf{ECE@k}} & {\textbf{Brier}}  \\
\midrule
 & \multicolumn{1}{l}{\textbf{Dataset}} & \multicolumn{2}{|c}{\textbf{LFAmazonTitles131K}} & \multicolumn{2}{|c}{\textbf{LFWikiSeeAlso320K}} \\ 
\midrule
 & \multicolumn{1}{l}{{GalaXC}} & \multicolumn{1}{|c}{28.78} & 21.77 & \multicolumn{1}{|c}{18.09} & 14.11 \\
& \multicolumn{1}{l}{{GalaXC-I}}  & \multicolumn{1}{|c}{0.04} & 12.3  & \multicolumn{1}{|c}{0.07} & 10.36  \\
 & \multicolumn{1}{l}{{NGAME}}  & \multicolumn{1}{|c}{19.29} & 15.74 & \multicolumn{1}{|c}{21.3} & 17.52  \\
& \multicolumn{1}{l}{{NGAME-I}}  & \multicolumn{1}{|c}{0.04} & 11.24 & \multicolumn{1}{|c}{0.09} & 12.43  \\
\midrule
 & \multicolumn{1}{l}{{Renee}}  & \multicolumn{1}{|c}{13.23} & 14.81 & \multicolumn{1}{|c}{12.29} & 13.14 \\
& \multicolumn{1}{l}{{Renee-I}}  & \multicolumn{1}{|c}{0.05} & 11.87 & \multicolumn{1}{|c}{0.06} & 11.38  \\
 & \multicolumn{1}{l}{{Gandalf}} & \multicolumn{1}{|c}{10.26} & 12.68 & \multicolumn{1}{|c}{12.15} & 14.58 \\
& \multicolumn{1}{l}{{Gandalf-I}}  & \multicolumn{1}{|c}{0.05} & 10.97  & \multicolumn{1}{|c}{0.05} & 11.91 \\

\bottomrule
\end{tabular}
\end{table}

% Even though label
% features are instrumental in obtaining boost in prediction performance (compared
% to approaches which ignore this information), it turns out that they are not
% particularly useful for calibration purposes. This is shown in Figure
% \ref{fig:fig1_experiment}, where it is demonstrated that the XMC methods which
% explicitly use label features, such as GalaXC and Ngame, do not exhibit any
% significant improvements over those (PLT and LightXML) which do not explicitly
% account for this additional information in the training data. In fact, the value
% of the calibration error metrics for GalaXC is relatively higher compared to the
% PLT approach.

\noindent \textbf{Meta-Classifier Strategies and Calibration:} Our empirical analysis reveals a significant correlation between meta-classifier strategies and model calibration in XMLC. The progression from randomized to sophisticated meta-classifiers consistently improves calibration metrics. LightXML \cite{jiang2021lightxml} with random meta-classifiers exhibits the highest ECE, while its original implementation with single-resolution, predefined meta-classifiers shows marked improvement. Notably, ELIAS \cite{gupta2022elias}, which learns label-to-cluster assignments concurrently with the training objective, demonstrates superior calibration across all k values. This indicates that adaptive meta-classifier construction may be crucial for well-calibrated predictions in these groups of methods. Interestingly, CascadeXML \cite{kharbanda2022cascadexml}, which employs a multi-resolution meta-classifier approach, shows poorer calibration compared to the single-resolution methods. This observation indicates that hierarchical strategies, while potentially enhancing precision, may not necessarily improve probability estimates across the label space.
We also note that the use of a meta-classifier in itself does not consistently lead to worse calibrations, even though it is unclear whether the resulting two-level scoring rule is still a proper loss. For example, Renee on Amazon-670K is less well calibrated than LightXML.

\noindent \textbf{ECE and ACE:}
Comparing fixed-size binning (ECE) and quantile binning (ACE) in \autoref{tab:main_table},
we see that generally, for the vanilla methods, the relative difference between
the two estimators is low. However, after recalibration, when the estimates are
much smaller overall, the differences between the two measures are of the same
order of magnitude as the estimates in themselves, so we cannot reliably rank
the recalibrated models; instead, all we can say is that they are much better
than the original models, and that they all achieve close to zero calibration 
error.

\begin{figure}
\centering
\includegraphics[width=\linewidth]{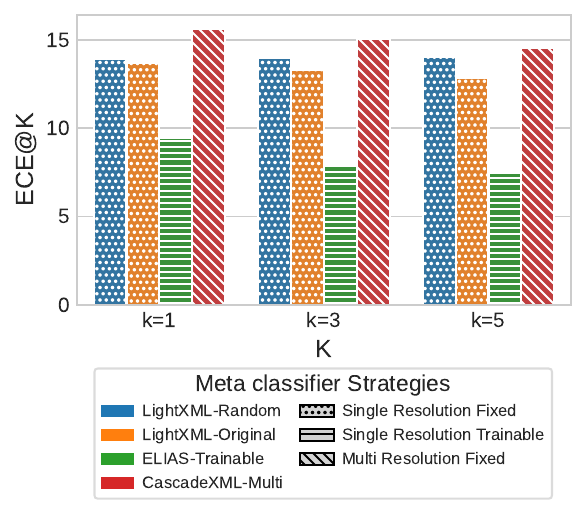} 
\caption{Impact of Meta Classifier Strategies on Calibration: Comparison of (i) fixed vs. trainable meta classifier assignment and (ii) single vs. multi-resolution meta classifiers.}
\label{fig:meta_experiments}
\end{figure}

\noindent \textbf{Scaling Effects and Additional Calibration Considerations:}
\autoref{fig:label_scaling} illustrates the effect of label scaling on the calibration performance of the Renee model. As the label size increases from 4K to 3M (millions), the ECE@$k$ rises, indicating a decline in calibration performance. This trend is anticipated, as the average number of data points per label diminishes with larger label sizes. Notably, the calibration error for AmazonCat-13K is lower compared to Eurlex-4K, attributable to the significantly higher average number of data points (\autoref{tab:dataset_statistics}) for AmazonCat-13K.

\noindent \emph{Joint vs. Separate Top-k Calibration:} We explored the impact of joint versus separate top-k calibration on ranking and precision metrics. However, no significant improvement was observed, as detailed in \autoref{sec:separate_k_calibration}.

\noindent \emph{Isotonic vs. Platt Scaling:} Our analysis revealed that isotonic regression consistently outperforms Platt scaling, with detailed results presented in \autoref{sec:post_hoc_comparison}.

%% file: parts/relatedwork.tex
%!TEX root = ../main.tex

\noindent{\textbf{Metrics for Assessing Predictive Model Calibration}}
Calibration misalignments are commonly quantified using the Expected Calibration Error ($\operatorname{ECE}$) \cite{naeini2015obtaining,guo2017calibration}, which measures the divergence between a model's predicted confidence and its empirical accuracy. While $\operatorname{ECE}$ is central to our calibration evaluation due to its widespread acceptance, accurately computing it presents challenges, primarily due to estimator bias—a topic extensively studied in recent literature ~\citep{NixonDZJT19,Roelofs2020,vaicenavicius_evaluating_2019,Gupta2021}.
To address calibration error bias, several approaches have been proposed: \citet{zhang2020mix} introduced a smoothed Kernel Density Estimation (KDE) method, \citet{brocker2012estimating}  and \citet{ferro2012bias} estimate and subtract per-bin bias, and \citet{Roelofs2020} employ equal mass bins, selecting the number of bins to preserve monotonicity in the calibration function.
For XMLC tasks, where standard $\operatorname{ECE}$ is less practical, we utilize the top-$k$ variant, denoted as $\operatorname{ECE@k}$. Our analysis also considers alternative metrics such as likelihood measures, the Brier score  \cite{Brier1950}, and Adaptive Calibration Error ($\operatorname{ACE}$)  \cite{NixonDZJT19} . These metrics and reliability diagrams \cite{DeGrootFienberg1983} are further analyzed in \autoref{sec:calib_matrics_vs_precision}.

\begin{figure}
\centering
\includegraphics[width=\linewidth]{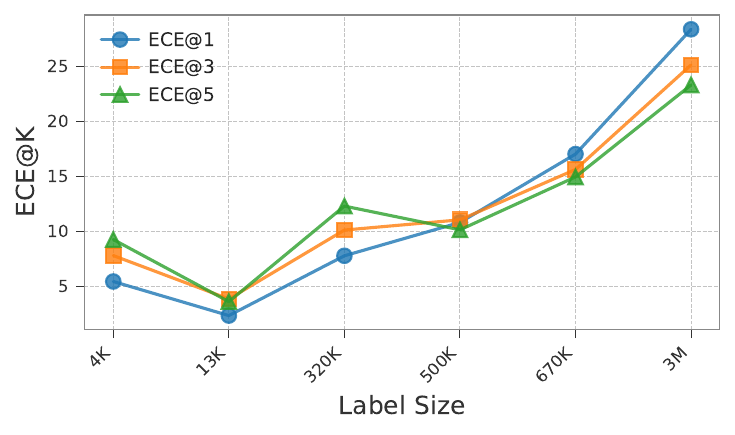} 
\caption{Impact of Calibration on Label Scaling: ECE@$K$ vs. Label Size for Renee Model.}
\label{fig:label_scaling}
\end{figure}

\noindent{\textbf{Empirical Analysis of model calibration.}} 
Recent empirical studies have focused on the calibration and robustness of models, illuminating significant findings. Notably, \citet{guo2017calibration} discovered that modern neural networks are less well-calibrated compared to those from the past decade, a trend more pronounced in larger networks which was observed to persist even as classification error decreased. Supporting evidence for these insights has been provided by works such as those by \citet{thulasidasan2019mixup}  and \citet{wen2020combining}, indicating a trend where larger models may compromise calibration, a consideration of paramount importance as the field progresses towards bigger models and datasets.
% In contrast, Winderer et al. have found that the calibration issues for recent models are inconsequential within the original data distribution and tend to improve when faced with new distributions. 
\citet{thulasidasan2019mixup} have further established that DNNs trained with the mixup technique are significantly better calibrated than those trained conventionally. Our research diverges from the established path, as we examine the calibration of XMLC models, especially in the face of statistical challenges like missing labels and the distribution of tail labels \cite{qaraei2021convex, schultheis2024generalized, schultheis2022missing}.

%% file: parts/appendix.tex
%!TEX root = ../main.tex
\clearpage

\section{XMLC Loss Functions: Calibration-Precision Trade-offs} 
\begin{figure}
\centering
\includegraphics[width=\columnwidth]{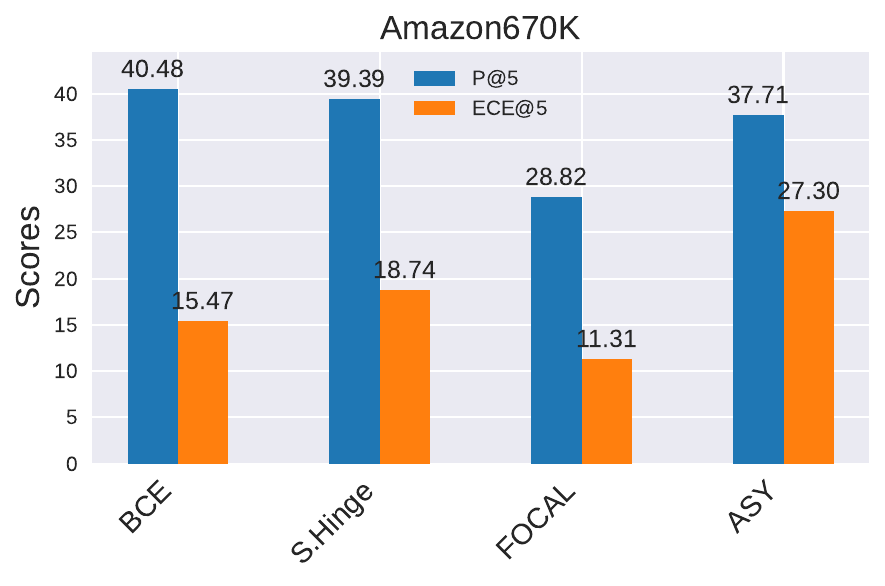} 
\caption{Calibration-Precision Comparison of XMLC Loss Functions (BCE, Squared Hinge, Focal, ASY) using Renee Model on Amazon-670K Dataset}
\label{fig:loss_plot}
\end{figure}

Figure \ref{fig:loss_plot} illustrates the calibration properties of popular loss functions in XMLC. We employed the Renee model \cite{jain2023renee} with fixed configurations, varying only the loss function to isolate its impact. Our analysis on the Amazon-670K dataset reveals a nuanced relationship between calibration and precision performance. The focal loss function, consistent with findings by  \citet{mukhoti2020calibrating}, achieves optimal calibration with an ECE@5 of 11.31\%. However, this comes at a significant cost to precision, yielding a P@5 of only 28.82\%. In contrast, the standard binary cross-entropy loss maintains a higher P@5 of 40.48\% while showing moderate calibration performance with an ECE@5 of 15.47\%. We also explored the ASY \cite{ridnik2021asymmetric} loss, popular in vision tasks for handling imbalanced label distributions and missing labels, previously unapplied in XMLC. Our results show that ASY \cite{ridnik2021asymmetric} loss not only fails to improve precision (P@5 of 37.71\%) but also significantly degrades calibration performance (ECE@5 of 27.30\%), aligning with observations by \citet{cheng2024towards}.

\begin{table}[ht]
\centering
\caption{Post-hoc calibration comparison on EurLex-4K and Amazon-670K Datasets. I and P denotes isotonic and platt scaling respectively.}
\label{tab:post_hoc_calibration_comparison}
\begin{tabular}{lcccccc}
\cline{1-7}
\multicolumn{1}{c}{} & \multicolumn{3}{c}{ECE@1 } & \multicolumn{3}{c}{ECE@5} \\
\hline
\multirow{2}{*}{Model} & \multicolumn{1}{c}{Pre-cal} & \multicolumn{2}{c}{Post-hoc} & \multicolumn{1}{c}{Pre-cal} & \multicolumn{2}{c}{Post-hoc} \\
\cline{3-4}\cline{6-7}
 & & I & P & & I & P \\
\hline
\multicolumn{7}{c}{EurLex-4K} \\
\hline
Dismec & 43.21 & \textit{0.75} & 1.85 & 32.86 & \textit{0.50} & 1.56 \\
PLT & 7.06 & \textit{1.19} & 2.30 & 9.67 & \textit{0.96} & 1.65 \\
AttentionXML & 8.31 & \textit{0.65} & 1.60 & 10.90 & \textit{0.32} & 5.86 \\
LightXML & 10.59 & 0.97 & \textit{1.37} & 13.28 & \textit{0.26} & 6.99 \\
CascadeXML & 10.19 & \textit{0.59} & 1.80 & 12.76 & \textit{0.43} & 2.71 \\
Renee & 5.43 & \textit{0.92} & 2.67 & 9.25 & \textit{0.59} & 2.68 \\
\hline
\multicolumn{7}{c}{Amazon-670K} \\
\hline
Dismec & 21.76 & \textit{0.11} & 2.41 & 17.76 & \textit{0.04} & 2.32 \\
PLT & 21.69 & \textit{0.09} & 4.47 & 19.70 & \textit{0.04} & 6.07 \\
AttentionXML & 16.83 & \textit{0.24} & 3.19 & 16.11 & \textit{0.20} & 8.07 \\
LightXML & 13.70 & \textit{0.15} & 5.18 & 12.86 & \textit{0.05} & 4.17 \\
CascadeXML & 15.63 & \textit{0.12} & 6.36 & 14.54 & \textit{0.04} & 4.68 \\
Renee & 17.02 & \textit{0.17} & 3.69 & 14.94 & \textit{0.05} & 5.19 \\
\hline
\end{tabular}
\end{table}

\section{Joint vs Separate Top-k Calibration}
\label{sec:separate_k_calibration}
We explore the implications of two calibration approaches in XMLC: joint calibration, which simultaneously adjusts probabilities for all top-$k$ labels while preserving their order, and separate calibration applied independently for each k. The latter may potentially alter the ranking of predictions post-calibration, consequently affecting performance metrics such as precision@k. Our analysis, utilizing both isotonic regression and Platt scaling across XMLC datasets, reveals that separate calibration generally results in slight performance degradation. On Amazon-670K, for example, CascadeXML's precision@1 decreases from 48.63\% to 48.32\% with separate isotonic calibration. Rare exceptions exist, such as LightXML on EurLex-4K, where precision@1 marginally improves from 85.90\% to 85.92\%. These observations suggest that preserving the original prediction order through joint calibration is generally advantageous for XMLC performance.

\section{Post-hoc Calibration: Isotonic vs Platt Scaling}
\label{sec:post_hoc_comparison}
Our empirical analysis, summarized in Table \ref{tab:post_hoc_calibration_comparison}, elucidates the comparative efficacy of isotonic regression and Platt scaling as post-hoc calibration methodologies for extreme multi-label classification (XMLC) tasks. Across six state-of-the-art XMLC algorithms and two benchmark datasets (Eurlex-4k and Amazon-670K), isotonic regression consistently demonstrates superior calibration performance, as quantified by Expected Calibration Error (ECE) at both top-1 and top-5 predictions. This performance disparity is more pronounced in the larger Amazon670K dataset, where isotonic regression achieves substantially lower ECE values compared to Platt scaling across all models. For instance, in the case of CascadeXML on Amazon-670K, isotonic regression yields an ECE@1 of 0.12 compared to 6.36 for Platt scaling. The non-parametric nature of isotonic regression enables it to capture complex relationships in the probability space of XMLC problems more effectively, with its performance advantage becoming more evident as the scale of the dataset increases. 

\section{Calibration Curves}
\begin{figure*}
\centering
\includegraphics[width=\textwidth]{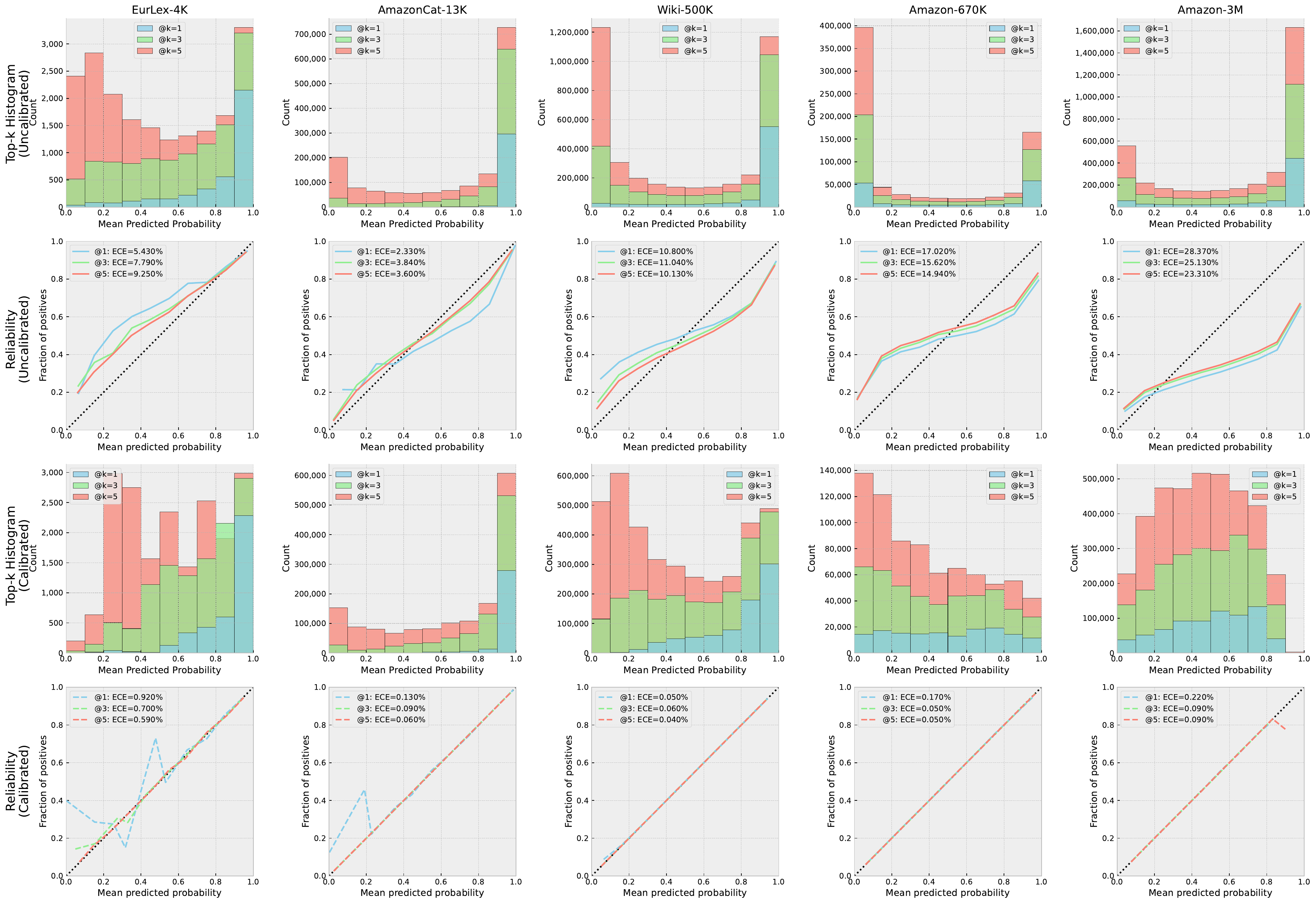} % Adjust the scale as necessary
\caption{Top-$k$ Calibration Curves and Histograms for \textit{Renee} across XMLC benchmarks. Top two rows: predicted probabilities and pre-calibration reliability. Bottom two rows: post-calibration metrics. Results for $k=1$, $k=3$, $k=5$, with ECE@$k$.}
\label{fig:fig2_renee_app}
\end{figure*}

\begin{figure*}
\centering
\includegraphics[width=\textwidth]{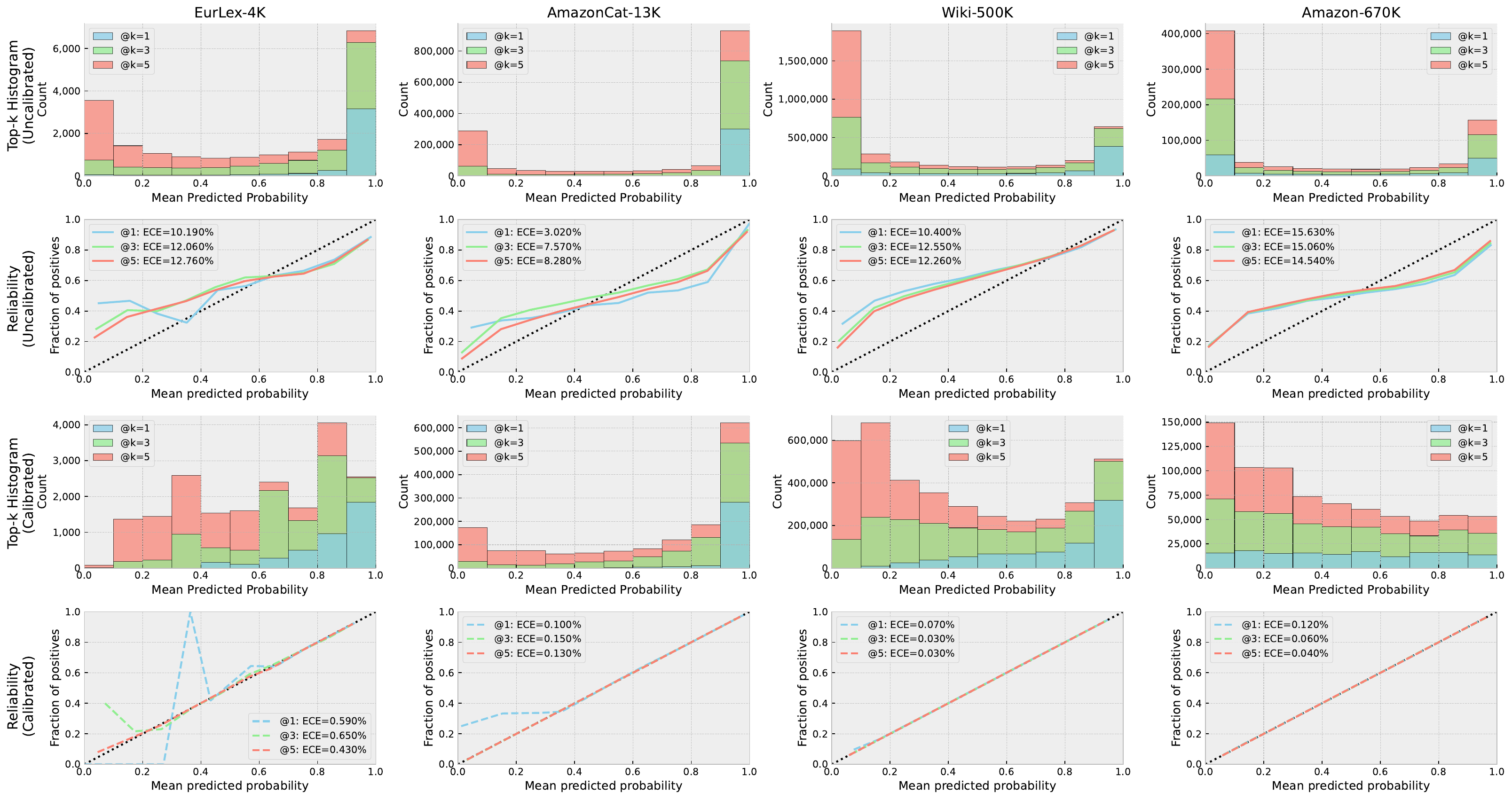} % Adjust the scale as necessary
\caption{Top-$k$ Calibration Curves and Histograms for \textit{CascadeXML} across XMLC benchmarks. Top two rows: predicted probabilities and pre-calibration reliability. Bottom two rows: post-calibration metrics. Results for $k=1$, $k=3$, $k=5$, with ECE@$k$.}
\label{fig:fig5_cascadexml_app}
\end{figure*}

\begin{figure*}
\centering
\includegraphics[width=\linewidth]{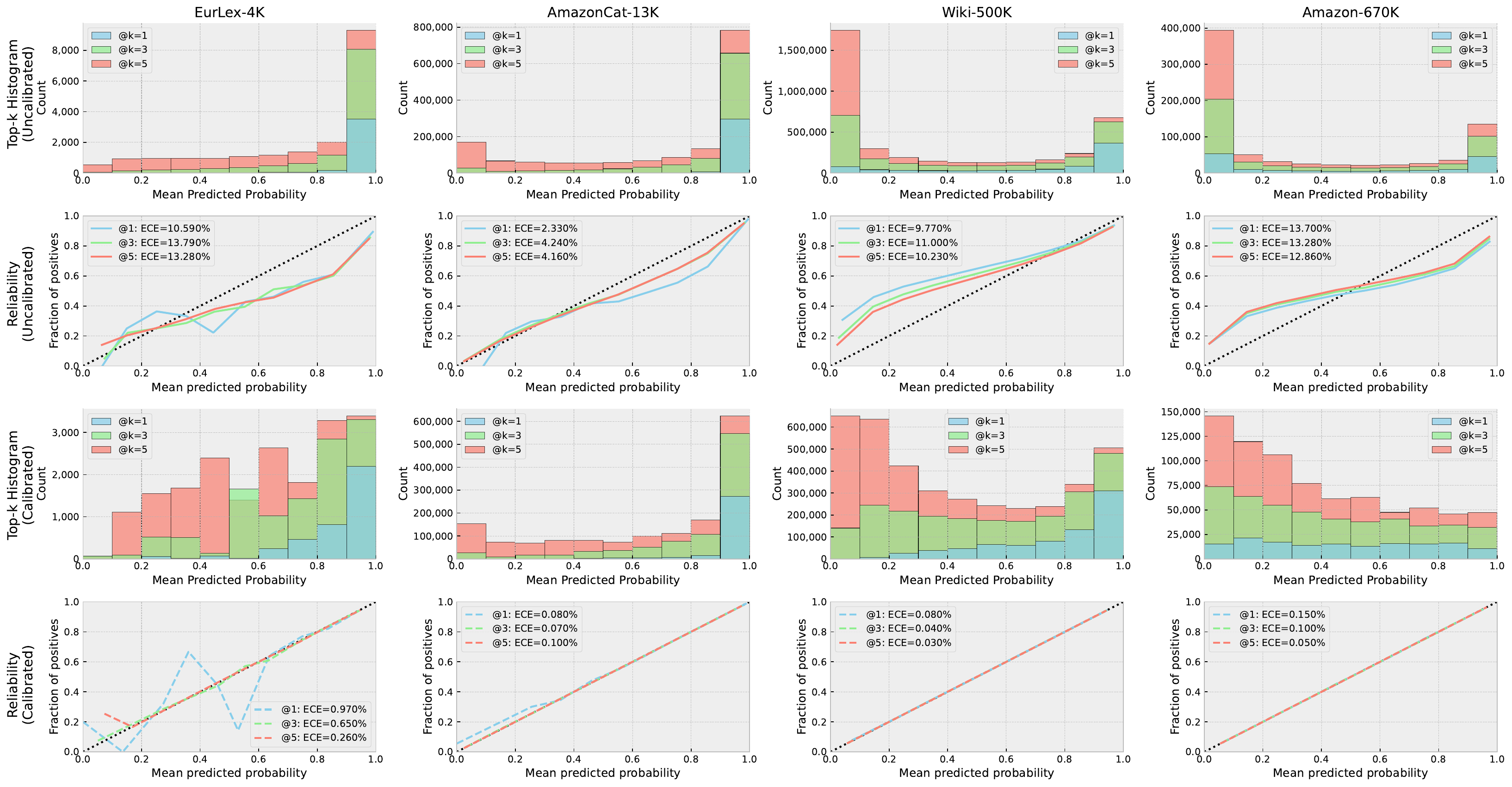} % Adjust the scale as necessary
\caption{Top-$k$ Calibration Curves and Histograms for \textit{LightXML} across XMLC benchmarks. Top two rows: predicted probabilities and pre-calibration reliability. Bottom two rows: post-calibration metrics. Results for $k=1$, $k=3$, $k=5$, with ECE@$k$.}
\label{fig:fig4_lightxml_app}
\end{figure*}

\begin{figure*}
\centering
\includegraphics[width=\linewidth]{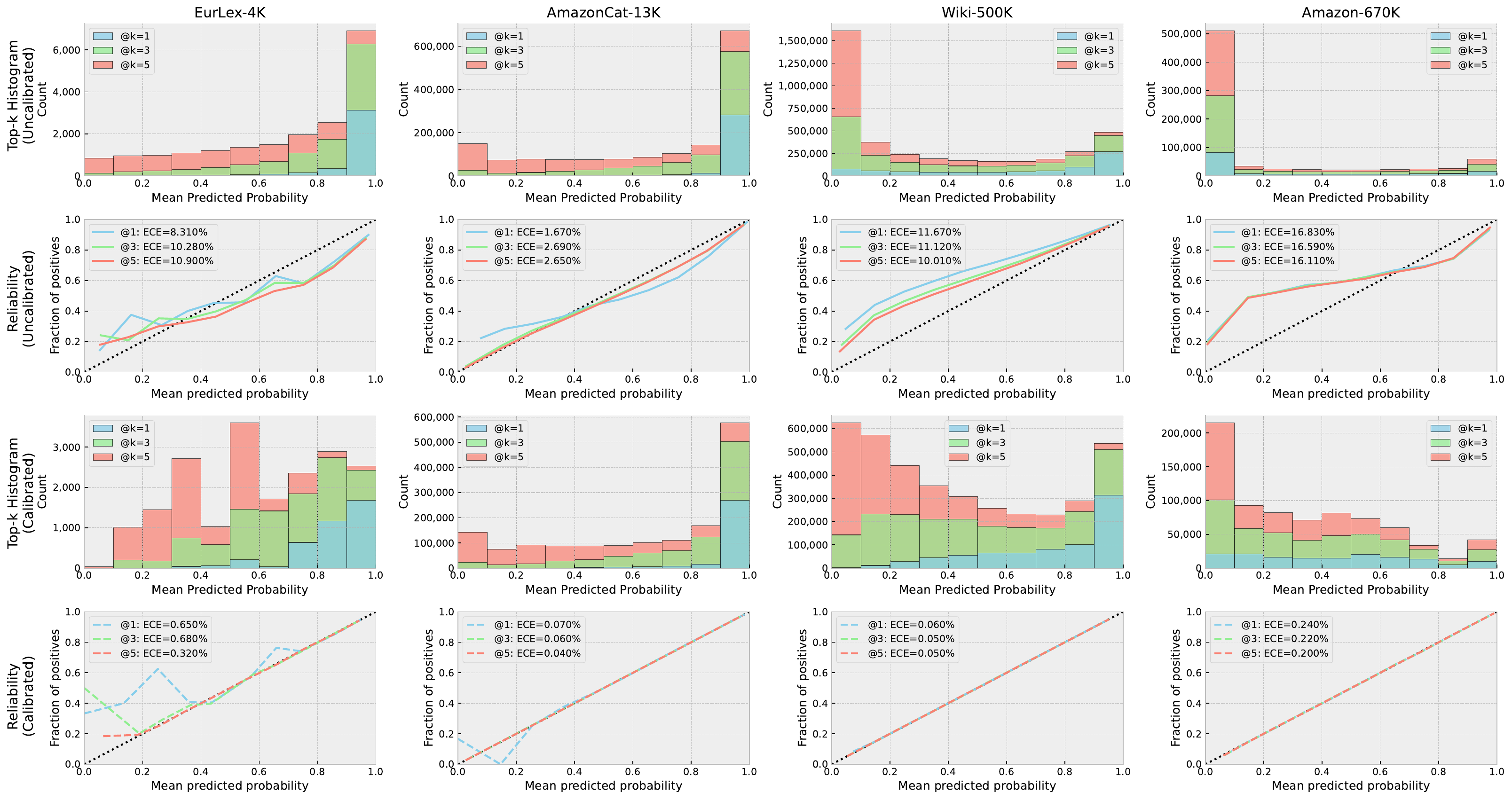} % Adjust the scale as necessary
\caption{Top-$k$ Calibration Curves and Histograms for \textit{AttentionXML} across XMLC benchmarks. Top two rows: predicted probabilities and pre-calibration reliability. Bottom two rows: post-calibration metrics. Results for $k=1$, $k=3$, $k=5$, with ECE@$k$.}
\label{fig:fig3_attentionxml_app}
\end{figure*}

\begin{figure*}
\centering
\includegraphics[width=\textwidth]{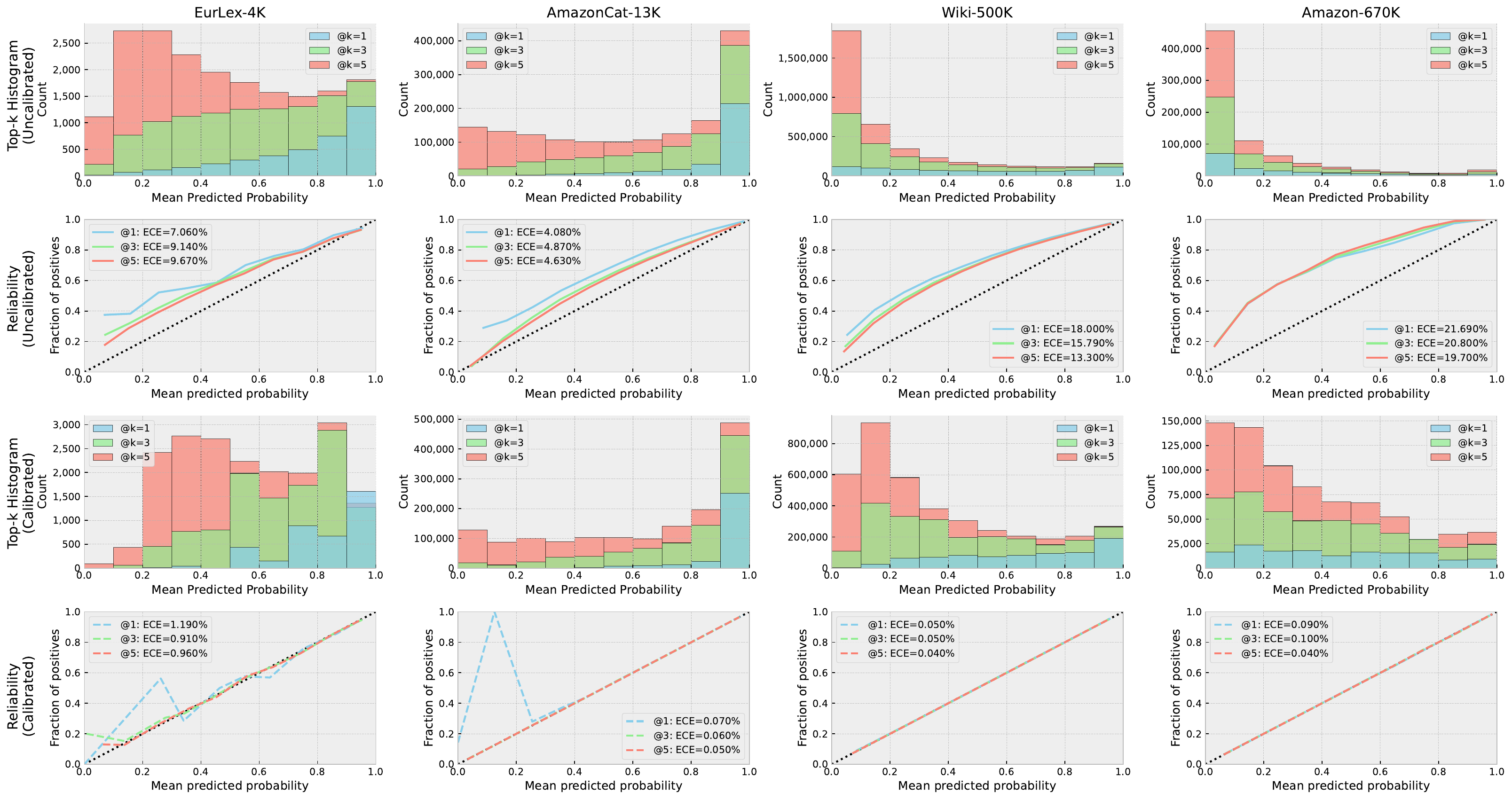} % Adjust the scale as necessary
\caption{Top-$k$ Calibration Curves and Histograms for \textit{PLT} across XMLC benchmarks. Top two rows: predicted probabilities and pre-calibration reliability. Bottom two rows: post-calibration metrics. Results for $k=1$, $k=3$, $k=5$, with ECE@$k$.}
\label{fig:fig2_plt_app}
\end{figure*}

\begin{figure*}
\centering
\includegraphics[width=\textwidth]{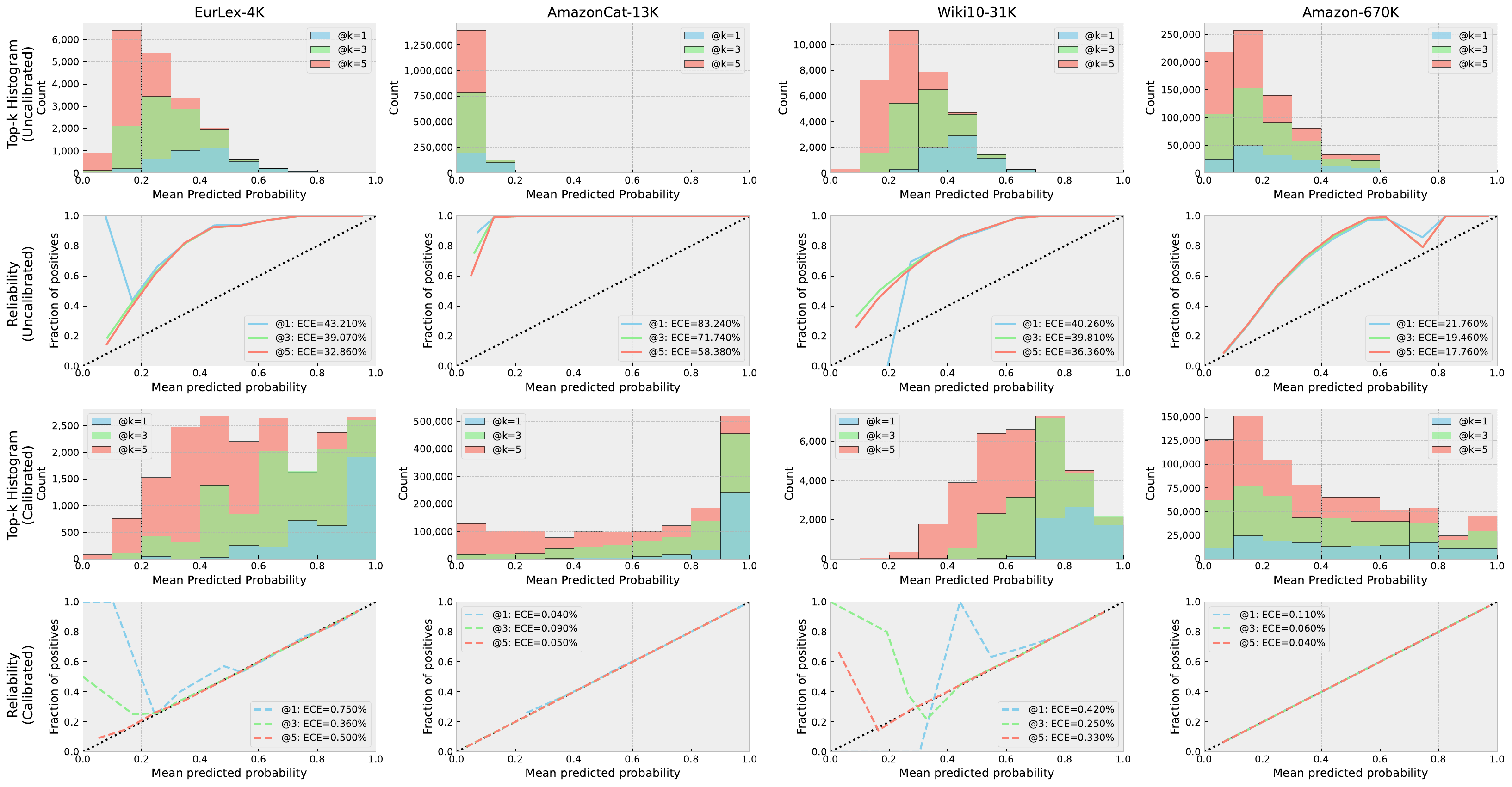} % Adjust the scale as necessary
\caption{Top-$k$ Calibration Curves and Histograms for \textit{Dismec} across XMLC benchmarks. Top two rows: predicted probabilities and pre-calibration reliability. Bottom two rows: post-calibration metrics. Results for $k=1$, $k=3$, $k=5$, with ECE@$k$.}
\label{fig:fig1_dismec_app}
\end{figure*}

\begin{figure*}
\includegraphics[width=\textwidth]{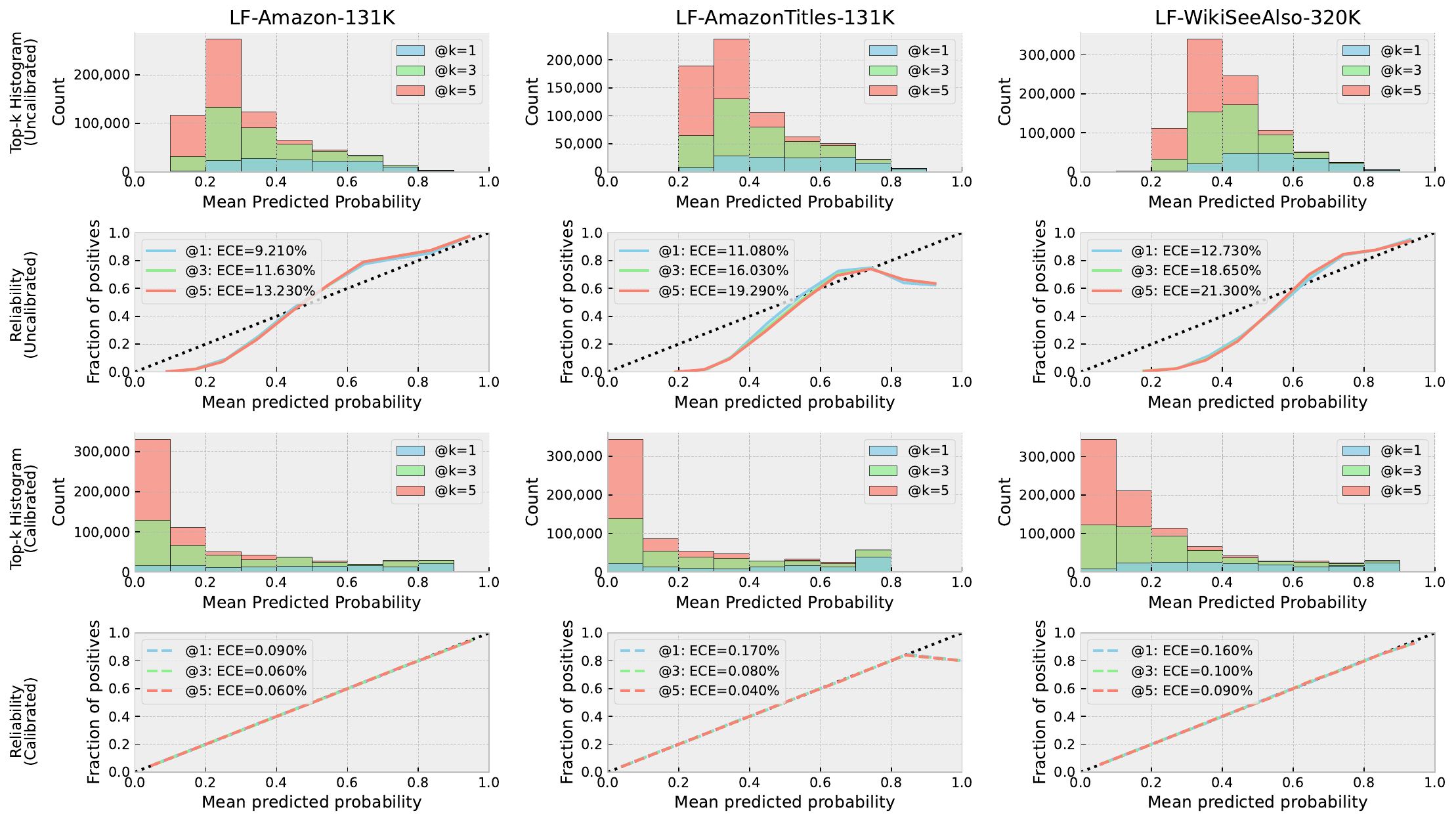} % Adjust the scale as necessary
\caption{Top-$k$ Calibration Curves and Histograms for \textit{NGAME} across Label feature (LF) based XMLC benchmarks. Top two rows: predicted probabilities and pre-calibration reliability. Bottom two rows: post-calibration metrics.}
\label{fig:fig1_ngame_app}
\end{figure*}

\begin{figure}[htbp]
\includegraphics[width=\linewidth]{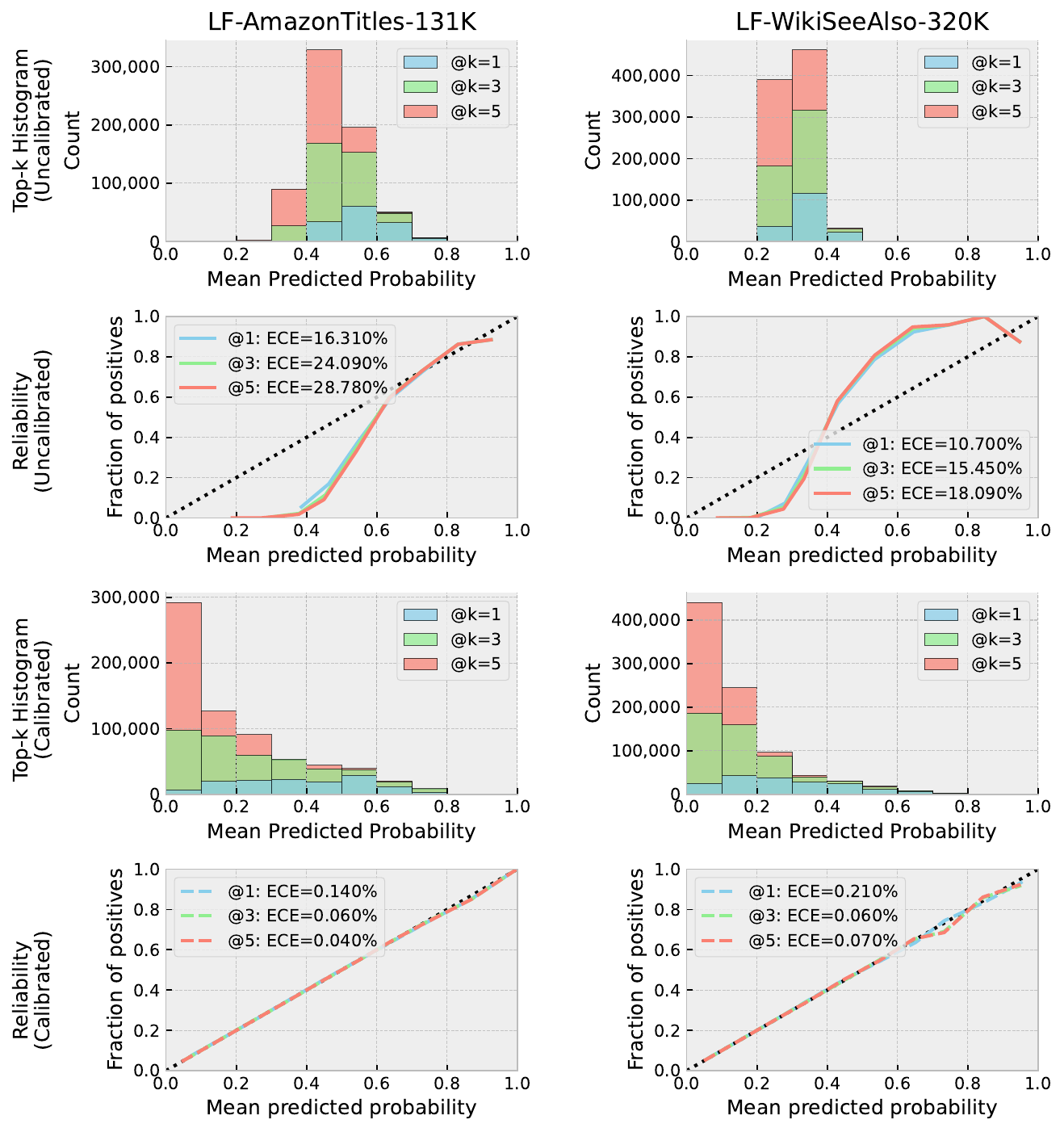} % Adjust the scale as necessary
\caption{Top-$k$ Calibration Curves and Histograms for \textit{GalaXC} across Label feature (LF) based XMLC benchmarks. Top two rows: predicted probabilities and pre-calibration reliability. Bottom two rows: post-calibration metrics.}
\label{fig:fig1_galaxc_app}
\end{figure}

\begin{figure}[htbp]
\includegraphics[width=\linewidth]{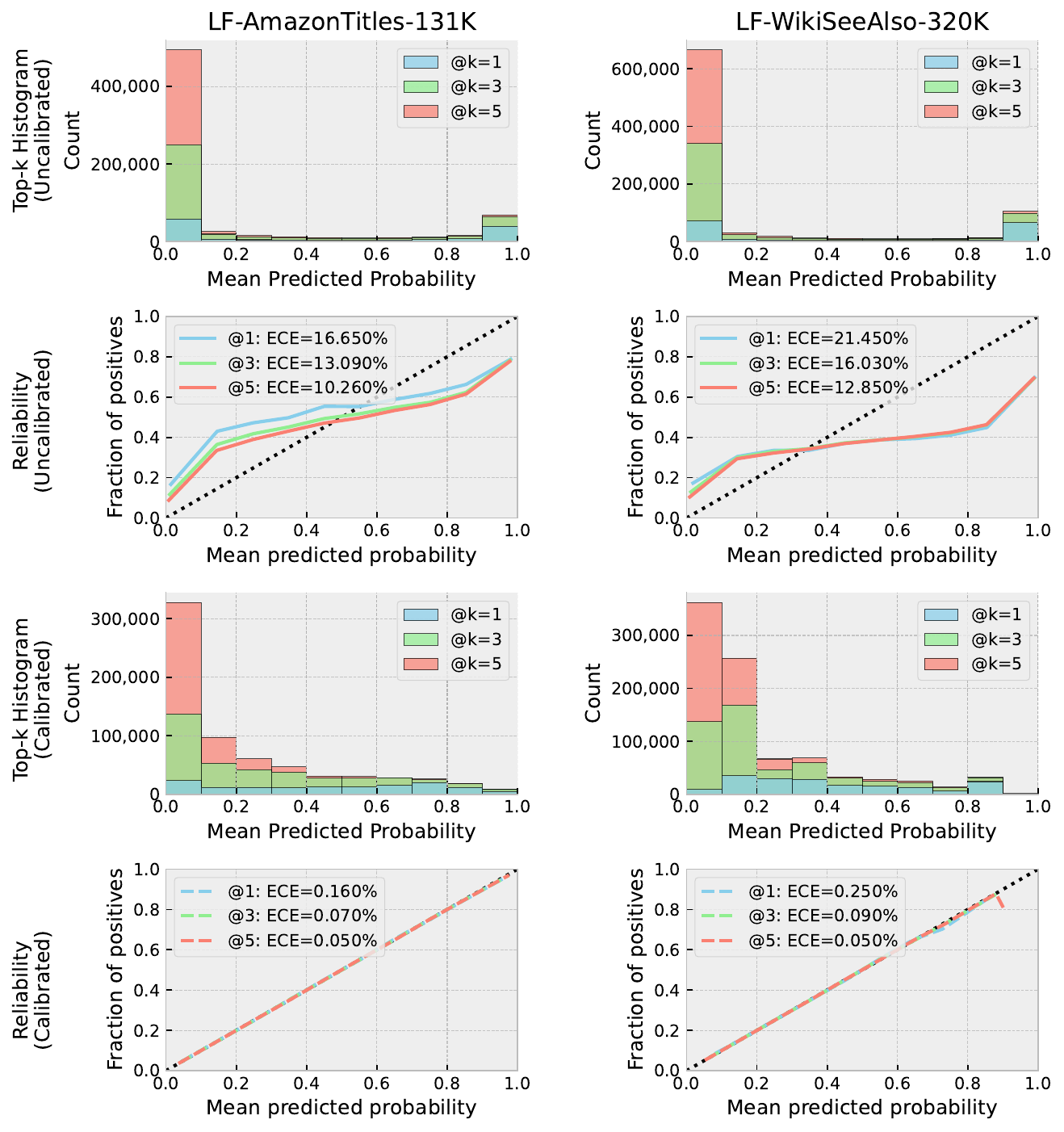} % Adjust the scale as necessary
\caption{Top-$k$ Calibration Curves and Histograms for \textit{GanDalf} \cite{kharbanda2024learning} across Label feature (LF) based XMLC benchmarks. Top two rows: predicted probabilities and pre-calibration reliability. Bottom two rows: post-calibration metrics.}
\label{fig:fig1_gandalf_app}
\end{figure}

\begin{figure}[htbp]
\includegraphics[width=\linewidth]{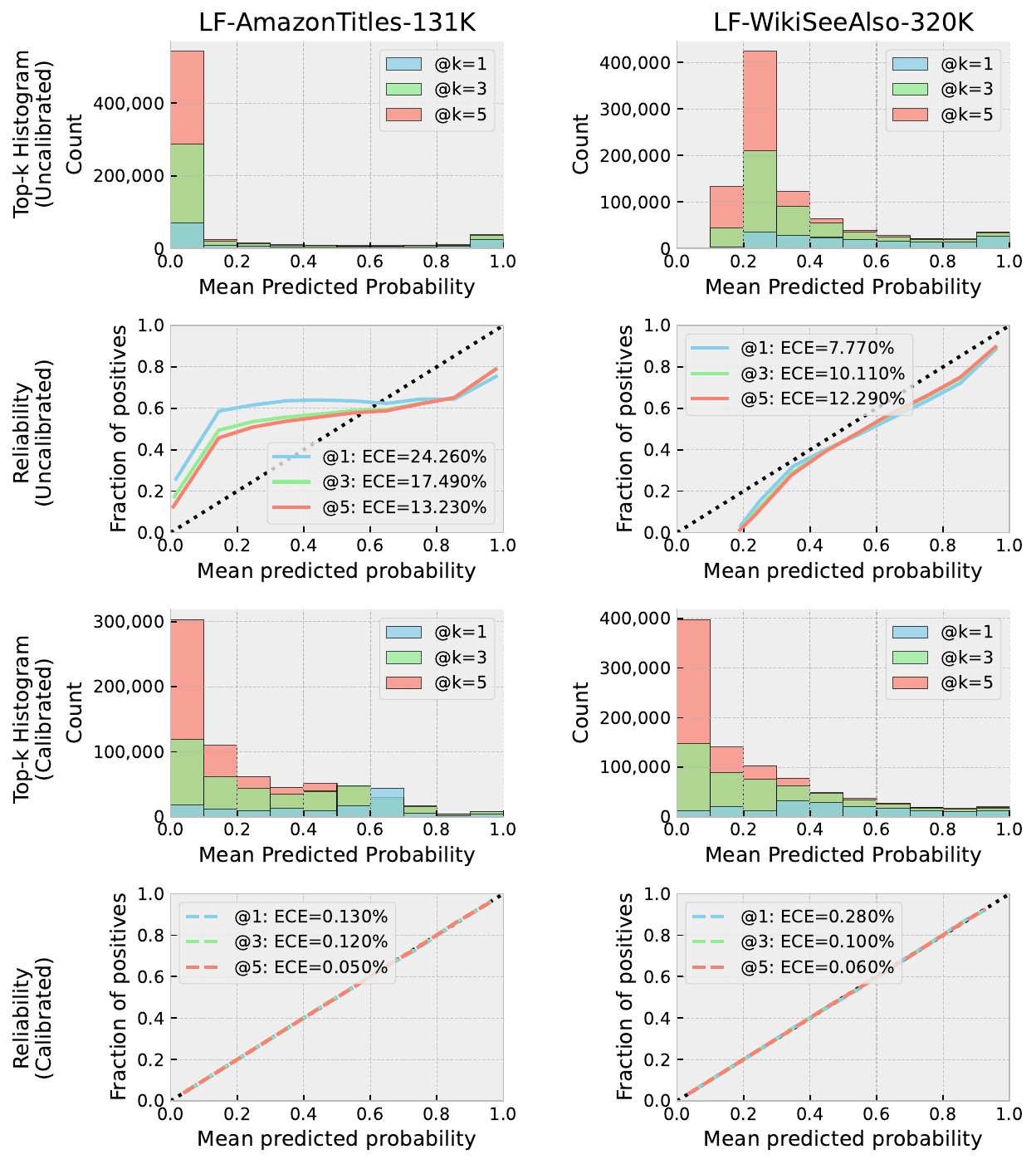} % Adjust the scale as necessary
\caption{Top-$k$ Calibration Curves and Histograms for \textit{Renee} across Label feature (LF) based XMLC benchmarks. Top two rows: predicted probabilities and pre-calibration reliability. Bottom two rows: post-calibration metrics.}
\label{fig:fig1_reneelf_app}
\end{figure}

Figures \ref{fig:fig2_renee_app}, \ref{fig:fig5_cascadexml_app}, \ref{fig:fig4_lightxml_app}, \ref{fig:fig3_attentionxml_app}, \ref{fig:fig2_plt_app}, and \ref{fig:fig1_dismec_app} illustrate the calibration curves for top-$k$ predictions of XMLC models Renee \cite{jain2023renee}, CascadeXML \cite{kharbanda2022cascadexml}, LightXML \cite{jiang2021lightxml}, AttentionXML \cite{You_et_al_2019}, PLT , and DiSMEC \cite{babbar2017dismec}, respectively. In each figure, the upper two rows depict the predicted top-$k$ probabilities and reliability diagrams for the original models, while the lower two rows present the same metrics for their post-hoc calibrated versions. For label-feature dataset methods, we present the calibration curves of NGAME \cite{NGAME}, GalaXC \cite{GalaXC}, GanDalf \cite{kharbanda2024learning}, and Renee \cite{jain2023renee} in Figures \ref{fig:fig1_ngame_app}, \ref{fig:fig1_galaxc_app}, \ref{fig:fig1_gandalf_app}, and \ref{fig:fig1_reneelf_app}, respectively.

\begin{table*}[ht]
\caption{Joint vs Separate Top-k Calibration on EurLex-4K and Amazon-670K. Calibrated-I and Calibrated-P indicates post-hoc calibration with isotonic and platt scaling respectively.}
\label{tab:combined_calibration}
\centering
\resizebox{\textwidth}{!}{%
\begin{tabular*}{\textwidth}{@{\extracolsep{\fill}}l*{9}{S[table-format=2.2]}}
\toprule
\multirow{2}{*}{\textbf{Model}} & \multicolumn{3}{c}{\textbf{P@1}} & \multicolumn{3}{c}{\textbf{P@3}} & \multicolumn{3}{c}{\textbf{P@5}} \\
\cmidrule(lr){2-4} \cmidrule(lr){5-7} \cmidrule(lr){8-10}
 & {Original} & {Calibrated-I} & {Calibrated-P} & {Original} & {Calibrated-I} & {Calibrated-P} & {Original} & {Calibrated-I} & {Calibrated-P} \\
\midrule
\multicolumn{10}{c}{\textbf{EurLex-4K}} \\
\midrule
Dismec & 83.75 & 83.54 & 83.75 & 70.71 & 70.53 & \bfseries 70.73 & 59.07 & 59.03 & 59.07 \\
PLT & 81.62 & 81.46 & 81.54 & 67.92 & 67.77 & 67.90 & 56.76 & 56.68 & 56.76 \\
AttentionXML & 84.63 & 83.83 & 84.63 & 72.73 & 72.48 & 72.63 & 60.66 & 60.61 & 60.66 \\
LightXML & 85.90 & \bfseries 85.92 & 85.85 & 74.02 & 72.14 & 73.92 & 62.23 & 62.14 & 62.17 \\
CascadeXML & 83.52 & 82.98 & 83.36 & 72.16 & 72.13 & 72.18 & 60.34 & 60.23 & 60.29 \\
\midrule
\multicolumn{10}{c}{\textbf{Amazon-670K}} \\
\midrule
Dismec & 45.66 & 45.43 & 45.39 & 40.66 & 40.55 & 40.55 & 37.18 & 37.13 & 37.14 \\
PLT & 44.22 & 44.11 & 44.03 & 39.29 & 39.23 & 39.24 & 35.68 & 35.67 & 35.67 \\
AttentionXML & 42.96 & 42.82 & 42.51 & 38.40 & 38.35 & 38.23 & 34.87 & 34.87 & 34.82 \\
LightXML & 47.29 & 46.81 & 46.02 & 42.31 & 42.09 & 41.84 & 38.51 & 38.46 & 38.39 \\
CascadeXML & 48.63 & 48.32 & 47.57 & 43.70 & 43.59 & 43.29 & 40.10 & 40.08 & 39.99 \\
\bottomrule
\end{tabular*}
}
\end{table*}

%Label Feature Methods

\clearpage

\section{Calibration Metrics in Relation to Precision}
\label{sec:calib_matrics_vs_precision}
\begin{figure*}
\centering
\includegraphics[width=\textwidth]{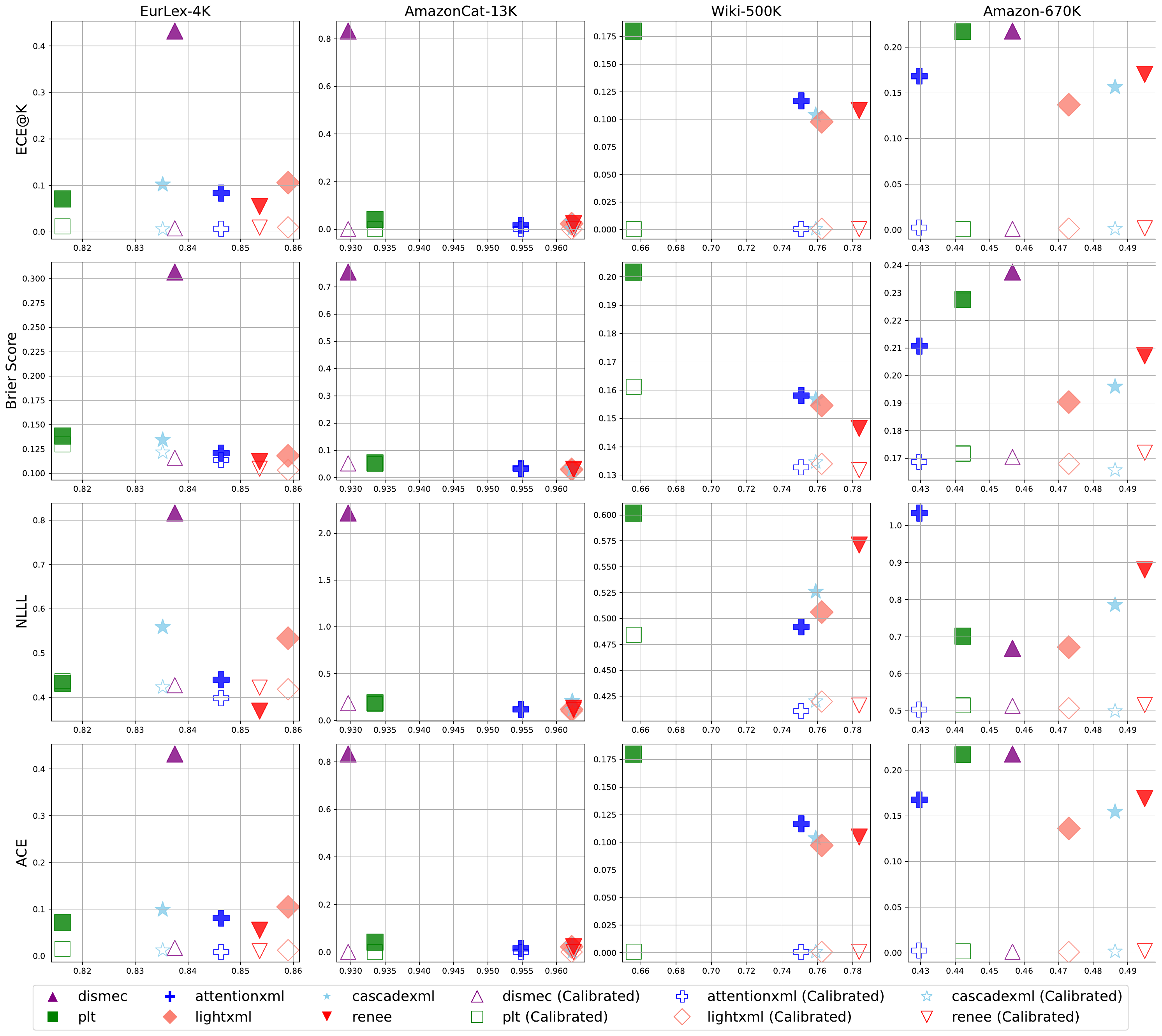} % Adjust the scale as necessary
\caption{Comparative visualization of calibration errors and performance metrics \textit{P@$1$} across multiple XMC datasets. Each column represents a different dataset and each row corresponds to a distinct metric evaluated. The first row illustrates Expected Calibration Error (ECE@$k$), followed by rows for Brier Score, Negative Log Likelihood (NLLL), and Adaptive Calibration Error (ACE), respectively. Solid markers indicate uncalibrated model scores, while hollow markers denote scores post-calibration. }
\label{fig:fig10_all_metrics_p1}
\end{figure*}

\begin{figure*}
\centering
\includegraphics[width=\textwidth]{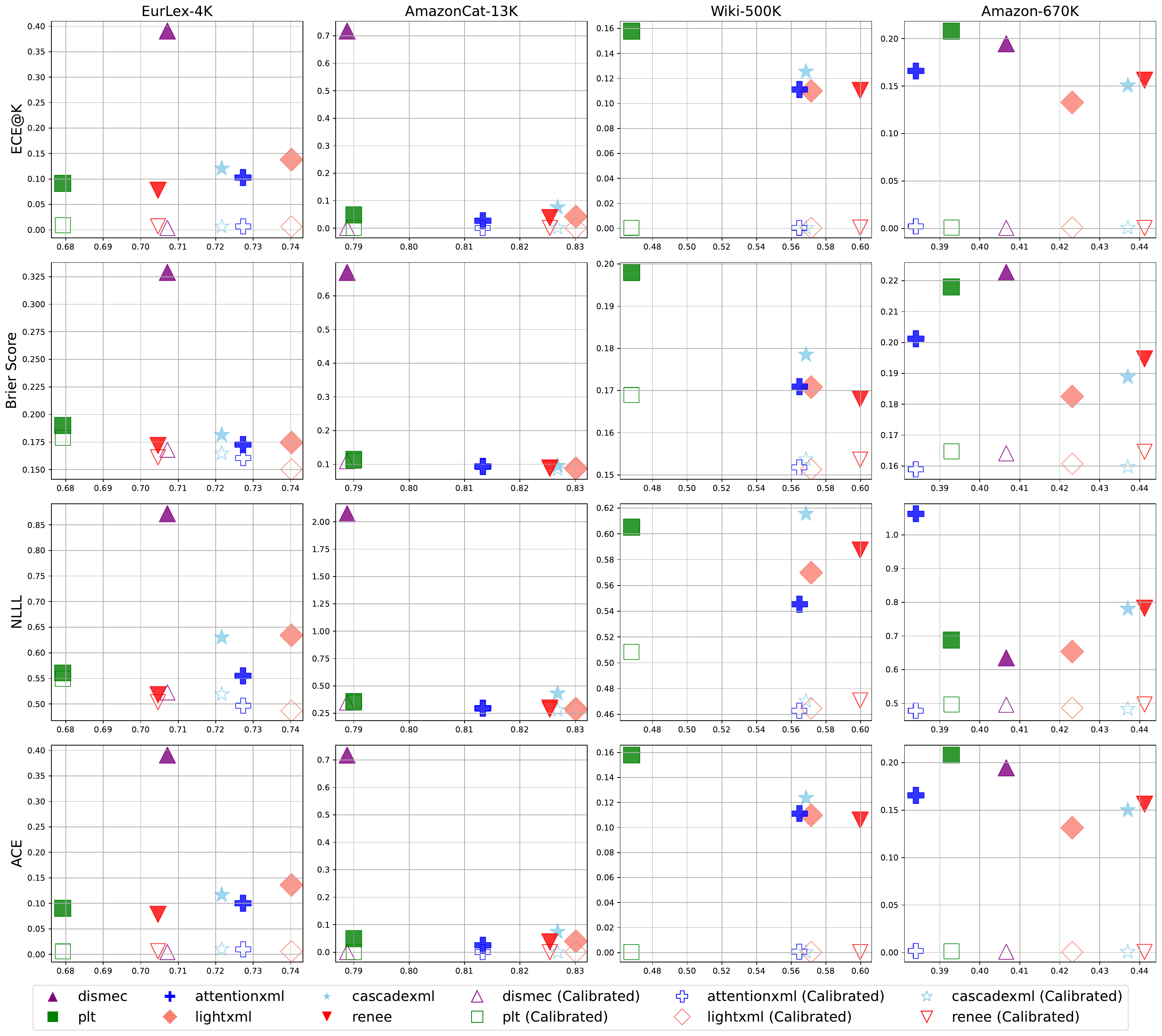} % Adjust the scale as necessary
\caption{Comparative visualization of calibration errors and performance metrics \textit{P@$3$} across multiple XMC datasets. Each column represents a different dataset and each row corresponds to a distinct metric evaluated. The first row illustrates Expected Calibration Error (ECE@$k$), followed by rows for Brier Score, Negative Log Likelihood (NLLL), and Adaptive Calibration Error (ACE), respectively. Solid markers indicate uncalibrated model scores, while hollow markers denote scores post-calibration. }
\label{fig:fig11_all_metrics_p3}
\end{figure*}

\begin{figure*}
\centering
\includegraphics[width=\textwidth]{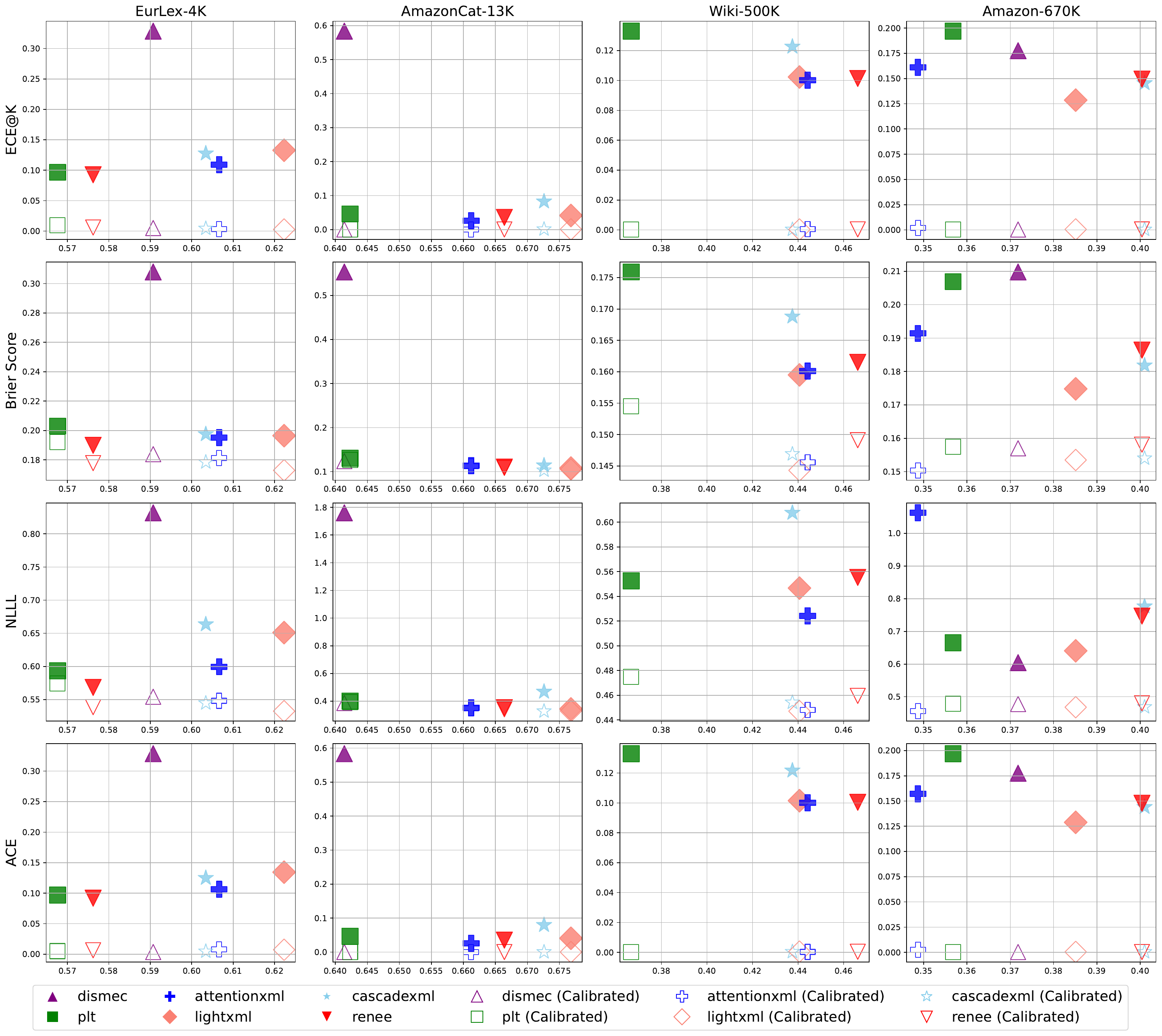} 
\caption{Comparative visualization of calibration errors and performance metrics \textit{P@$5$} across multiple XMC datasets. Each column represents a different dataset and each row corresponds to a distinct metric evaluated. The first row illustrates Expected Calibration Error (ECE@$k$), followed by rows for Brier Score, Negative Log Likelihood (NLLL), and Adaptive Calibration Error (ACE), respectively. Solid markers indicate uncalibrated model scores, while hollow markers denote scores post-calibration.}
\label{fig:fig12_all_metrics_p5}
\end{figure*}

 \autoref{fig:fig10_all_metrics_p1}, \ref{fig:fig11_all_metrics_p3}, and \ref{fig:fig12_all_metrics_p5} provide a comprehensive comparison of calibration errors in relation to Precision@$k$ across multiple XMLC datasets. Each figure represents different levels of Precision@k: P@1, P@3, and P@5, respectively. The datasets evaluated include EurLex-4K, AmazonCat-13K, Wiki-500K, and Amazon-670K, with each column in the figures corresponding to one of these datasets. The rows in the figures depict various calibration metrics: Expected Calibration Error (ECE@$k$) Brier Score, Negative Log Likelihood (NLL), and Adaptive Calibration Error (ACE). Solid markers indicate uncalibrated model scores, while hollow markers denote scores post-calibration.

\section{Calibration Performance of XMLC Algorithms:}
% \label{sec:extended_main_table}
\autoref{tab:main_table_extended} presents comprehensive quantitative measurements of various calibration metrics for non-label-feature methods and datasets. Correspondingly, \autoref{tab:LF_table_extended} provides analogous metrics for label-feature methods and datasets.

\begin{table*}
\centering
\caption{Calibration error metrics on four datasets for a range of XMLC algorithms. Two consecutive rows, such as DiSMEC and DiSMEC-I, represent the measures on the vanilla version of the algorithm and that obtained by re-calibration via Isotonic regression respectively. It can be observed that the calibration error can be significantly reduced by this methodology.}
\label{tab:main_table_extended}
\begin{tabular}{
  @{}
  l
  l
  S[table-format=2.2]
  S[table-format=2.2]
  S[table-format=2.2]
  S[table-format=2.2]
  S[table-format=2.2]
  S[table-format=2.2]
  S[table-format=2.2]
  S[table-format=2.2]
  S[table-format=2.2]
  S[table-format=2.2]
  S[table-format=2.2]
  S[table-format=2.2]
  @{}
}
\toprule
 & \textbf{Calibration Measure} &  \multicolumn{1}{|c}{\textbf{ECE-k}} & {\textbf{Brier}} & {\textbf{ACE}} & \multicolumn{1}{|c}{\textbf{ECE-k}} & {\textbf{Brier}} & {\textbf{ACE}} & \multicolumn{1}{|c}{\textbf{ECE-k}} & {\textbf{Brier}} & {\textbf{ACE}} & \multicolumn{1}{|c}{\textbf{ECE-k}}& {\textbf{Brier}} & {\textbf{ACE}} \\
\midrule
 & \multicolumn{1}{c}{\textbf{Dataset}} & \multicolumn{3}{|c}{\textbf{Eurlex-4K}} & \multicolumn{3}{|c}{\textbf{AmazonCat-13K}} & \multicolumn{3}{|c}{\textbf{Wiki-500K}} & \multicolumn{3}{|c}{\textbf{Amazon-670K}} \\ 
\midrule
\multirow{10}{*}{k=1} & \multicolumn{1}{|c}{{DiSMEC}} & \multicolumn{1}{|c}{43.21} & 30.69 & 43.21 &  \multicolumn{1}{|c}{83.24} & 75.43 & 83.24 & \multicolumn{1}{|c}{-} & \multicolumn{1}{c}{-} & \multicolumn{1}{c}{-} & \multicolumn{1}{|c}{21.76} & 23.76 & 21.76 \\
& \multicolumn{1}{|c}{{DiSMEC-I}}  & \multicolumn{1}{|c}{0.75} & 1.60 & 1.73  & \multicolumn{1}{|c}{0.04} & 0.06 & 18.45 & \multicolumn{1}{|c}{-} & \multicolumn{1}{c}{-} & \multicolumn{1}{c}{-} & \multicolumn{1}{|c}{0.11} & 17.04 & 0.13 \\
& \multicolumn{1}{|c}{{PLT}}  & \multicolumn{1}{|c}{7.06} & 13.83 & 7.13  & \multicolumn{1}{|c}{4.08} & 5.41 & 4.14 & \multicolumn{1}{|c}{18.0} & 20.17 & 18.00 & \multicolumn{1}{|c}{21.69} &22.76 &21.69 \\
& \multicolumn{1}{|c}{{PLT-I}}  & \multicolumn{1}{|c}{1.19} & 12.98 & 1.47  & \multicolumn{1}{|c}{0.07} & 4.94 & 0.05 & \multicolumn{1}{|c}{0.05} & 16.12 & 0.08 & \multicolumn{1}{|c}{0.09} & 17.17 & 0.16 \\
& \multicolumn{1}{|c}{{AttentionXML}}  & \multicolumn{1}{|c}{8.31} & 12.09 & 8.08  & \multicolumn{1}{|c}{1.67} & 3.36 & 1.6 & \multicolumn{1}{|c}{11.67} & 15.81 & 11.67 & \multicolumn{1}{|c}{16.83} & 21.07 & 16.75 \\
& \multicolumn{1}{|c}{{AttentionXML-I}}  & \multicolumn{1}{|c}{0.65} & 11.37 & 0.83  & \multicolumn{1}{|c}{0.07} & 3.24 & 0.02 & \multicolumn{1}{|c}{0.06} & 13.27 & 0.06 & \multicolumn{1}{|c}{0.24} & 16.86 & 0.24\\
& \multicolumn{1}{|c}{{LightXML}}  & \multicolumn{1}{|c}{10.59}& 11.81 & 10.50  & \multicolumn{1}{|c}{2.33} & 3.03 & 2.26 & \multicolumn{1}{|c}{9.77} & 15.46 & 9.72 & \multicolumn{1}{|c}{13.7} & 19.04 &13.62 \\
& \multicolumn{1}{|c}{{LightXML-I}}  & \multicolumn{1}{|c}{0.97}& 10.34 & 1.23  & \multicolumn{1}{|c}{0.08} & 2.77& 0.03 & \multicolumn{1}{|c}{0.08} & 13.4 & 0.07 & \multicolumn{1}{|c}{0.15} & 16.8 & 0.09 \\
& \multicolumn{1}{|c}{{CascadeXML}}  & \multicolumn{1}{|c}{10.19} & 13.44 & 9.9  & \multicolumn{1}{|c}{3.02} & 3.26 & 2.86 & \multicolumn{1}{|c}{10.4} & 15.68 & 10.38 & \multicolumn{1}{|c}{15.63} & 19.6 & 15.44 \\
& \multicolumn{1}{|c}{{CascadeXML-I}}  & \multicolumn{1}{|c}{0.59} & 12.17 & 1.23 & \multicolumn{1}{|c}{0.1} & 2.88 & 0.03 & \multicolumn{1}{|c}{0.07} & 13.46 & 0.08 & \multicolumn{1}{|c}{0.12} & 16.57 & 0.16 \\
& \multicolumn{1}{|c}{{Renee}}  & \multicolumn{1}{|c}{5.43}& 11.21 & 5.5  & \multicolumn{1}{|c}{2.33} & 3.05 & 2.24 & \multicolumn{1}{|c}{10.8} & 14.65 & 10.48 & \multicolumn{1}{|c}{17.02} & 20.72 & 16.88 \\
& \multicolumn{1}{|c}{{Renee-I}}  & \multicolumn{1}{|c}{0.92}& 10.48 & 1.04  & \multicolumn{1}{|c}{0.13} & 2.85 & 0.04 & \multicolumn{1}{|c}{0.05} & 13.18 & 0.08 & \multicolumn{1}{|c}{0.17} & 17.2 & 0.21 \\
\midrule
\multirow{10}{*}{k=3} & \multicolumn{1}{|c}{{DiSMEC}} & \multicolumn{1}{|c}{39.07} & 32.89 & 39.07  & \multicolumn{1}{|c}{71.74} & 66.97 &71.74 & \multicolumn{1}{|c}{-} & \multicolumn{1}{c}{-} & \multicolumn{1}{c}{-} & \multicolumn{1}{|c}{19.46} & 22.27 & 19.46 \\
& \multicolumn{1}{|c}{{DiSMEC-I}}  & \multicolumn{1}{|c}{0.36} & 1.68& 0.43 & \multicolumn{1}{|c}{0.09} & 10.92& 0.03 & \multicolumn{1}{|c}{-} & \multicolumn{1}{c}{-} & \multicolumn{1}{c}{-} & \multicolumn{1}{|c}{0.06} & 16.41& 0.08 \\
& \multicolumn{1}{|c}{{PLT}}  & \multicolumn{1}{|c}{9.14} & 19.01 & 9.05  & \multicolumn{1}{|c}{4.87} &11.46& 4.91 & \multicolumn{1}{|c}{15.79} & 19.8 & 15.79 & \multicolumn{1}{|c}{20.8} &21.8 &20.8 \\
& \multicolumn{1}{|c}{{PLT-I}}  & \multicolumn{1}{|c}{0.91} & 1.79& 0.55  & \multicolumn{1}{|c}{0.06} & 11.01 & 0.03 & \multicolumn{1}{|c}{0.05} & 16.89 & 0.05 & \multicolumn{1}{|c}{0.1} & 16.47 & 0.11 \\
& \multicolumn{1}{|c}{{AttentionXML}}  & \multicolumn{1}{|c}{10.28} & 17.28 &10.0  & \multicolumn{1}{|c}{2.69} & 9.38 & 2.52 & \multicolumn{1}{|c}{11.12} & 17.09 & 11.12 & \multicolumn{1}{|c}{16.59} & 20.12 & 16.57 \\
& \multicolumn{1}{|c}{{AttentionXML-I}}  & \multicolumn{1}{|c}{0.68} & 16.05 & 0.96  & \multicolumn{1}{|c}{0.06} & 9.26 & 0.02 & \multicolumn{1}{|c}{0.05} & 15.18 & 0.07 & \multicolumn{1}{|c}{0.22} & 15.89 & 0.18\\
& \multicolumn{1}{|c}{{LightXML}}  & \multicolumn{1}{|c}{13.79}& 17.46 & 13.61  & \multicolumn{1}{|c}{4.24} & 8.81 & 3.97 & \multicolumn{1}{|c}{11.0} & 17.08 & 10.98 & \multicolumn{1}{|c}{13.28} & 18.25 &13.15 \\
& \multicolumn{1}{|c}{{LightXML-I}}  & \multicolumn{1}{|c}{0.65}& 15.02 & 0.58  & \multicolumn{1}{|c}{0.07} & 8.49& 0.02 & \multicolumn{1}{|c}{0.04} & 15.13 & 0.04 & \multicolumn{1}{|c}{0.1} & 16.07 & 0.04 \\
& \multicolumn{1}{|c}{{CascadeXML}}  & \multicolumn{1}{|c}{12.06} & 18.15 & 11.64  & \multicolumn{1}{|c}{7.57} & 9.54 & 7.43 & \multicolumn{1}{|c}{12.55} & 17.85 & 12.37 & \multicolumn{1}{|c}{15.06} &18.88 & 15.0 \\
& \multicolumn{1}{|c}{{CascadeXML-I}}  & \multicolumn{1}{|c}{0.65} & 16.47 & 1.0 & \multicolumn{1}{|c}{0.15} & 8.51 & 0.03 & \multicolumn{1}{|c}{0.03} & 15.37 & 0.05 & \multicolumn{1}{|c}{0.06} & 15.97 & 0.06 \\
& \multicolumn{1}{|c}{{Renee}}  & \multicolumn{1}{|c}{7.79}& 17.19 & 7.84  & \multicolumn{1}{|c}{3.84} & 8.92 & 3.74 & \multicolumn{1}{|c}{11.04} & 16.8 & 10.6 & \multicolumn{1}{|c}{15.62} & 19.47 & 15.63 \\
& \multicolumn{1}{|c}{{Renee-I}}  & \multicolumn{1}{|c}{0.7}& 16.11 & 0.62  & \multicolumn{1}{|c}{0.09} & 8.69 & 0.02 & \multicolumn{1}{|c}{0.06} & 15.36 & 0.04 & \multicolumn{1}{|c}{0.05} & 16.46 & 0.06 \\
\midrule
\multirow{10}{*}{k=5} & \multicolumn{1}{|c}{{DiSMEC}} & \multicolumn{1}{|c}{32.86} & 30.79 & 32.86  & \multicolumn{1}{|c}{58.38} & 55.32 & 58.38 & \multicolumn{1}{|c}{-} & \multicolumn{1}{c}{-}  & \multicolumn{1}{c}{-}  & \multicolumn{1}{|c}{17.76} & 20.98 & 17.76 \\
& \multicolumn{1}{|c}{{DiSMEC-I}}  & \multicolumn{1}{|c}{0.05} & 18.39 & 0.36  & \multicolumn{1}{|c}{0.05} & 1.24 & 0.03 & \multicolumn{1}{|c}{-} & \multicolumn{1}{c}{-} & \multicolumn{1}{c}{-} & \multicolumn{1}{|c}{0.04} & 1.57 & 0.03 \\
& \multicolumn{1}{|c}{{PLT}}  & \multicolumn{1}{|c}{9.67} & 2.03 & 9.71  & \multicolumn{1}{|c}{4.63} & 13.05 & 4.65 & \multicolumn{1}{|c}{13.3} & 17.59 & 13.3 & \multicolumn{1}{|c}{19.7} & 20.69 & 19.7 \\
& \multicolumn{1}{|c}{{PLT-I}}  & \multicolumn{1}{|c}{0.96} & 19.22& 0.51  & \multicolumn{1}{|c}{0.05} & 12.68 & 0.02 & \multicolumn{1}{|c}{0.04} & 15.45 & 0.02 & \multicolumn{1}{|c}{0.04} & 15.75 & 0.04 \\
& \multicolumn{1}{|c}{{AttentionXML}}  & \multicolumn{1}{|c}{10.9} & 19.51 & 10.63 & \multicolumn{1}{|c}{2.65} & 11.36 & 2.59 & \multicolumn{1}{|c}{10.01} & 16.01 & 10.01 & \multicolumn{1}{|c}{16.11} & 19.14 & 15.71 \\
& \multicolumn{1}{|c}{{AttentionXML-I}}  & \multicolumn{1}{|c}{0.32} &18.12 & 0.78  & \multicolumn{1}{|c}{0.04} & 11.24 & 0.02 & \multicolumn{1}{|c}{0.05} & 14.56 & 0.03 & \multicolumn{1}{|c}{0.02} & 15.04 & 0.26\\
& \multicolumn{1}{|c}{{LightXML}}  & \multicolumn{1}{|c}{13.28}& 19.65 & 13.44 & \multicolumn{1}{|c}{4.16} & 10.83 & 4.07 & \multicolumn{1}{|c}{10.23} & 15.95 & 10.15 & \multicolumn{1}{|c}{12.86} & 17.48 & 12.89 \\
& \multicolumn{1}{|c}{{LightXML-I}}  & \multicolumn{1}{|c}{0.26} & 17.29 &0.74  & \multicolumn{1}{|c}{0.01} & 10.52 & 0.02 & \multicolumn{1}{|c}{0.03} & 14.43 & 0.05 & \multicolumn{1}{|c}{0.05} & 15.35 & 0.04 \\
& \multicolumn{1}{|c}{{CascadeXML}}  & \multicolumn{1}{|c}{12.76} & 19.75 & 12.5  & \multicolumn{1}{|c}{8.28} & 11.43& 9.79 & \multicolumn{1}{|c}{12.26} & 16.88 & 12.17 & \multicolumn{1}{|c}{14.54} & 18.18 & 14.41 \\
& \multicolumn{1}{|c}{{CascadeXML-I}}  & \multicolumn{1}{|c}{0.43} & 17.85 & 0.47 & \multicolumn{1}{|c}{0.13} & 10.31 & 0.03  & \multicolumn{1}{|c}{0.03} & 14.69 & 0.04 & \multicolumn{1}{|c}{0.04} & 15.4 & 0.04 \\
& \multicolumn{1}{|c}{{Renee}}  & \multicolumn{1}{|c}{9.25}& 18.99 & 9.18 & \multicolumn{1}{|c}{3.6} & 11.05 & 3.52  & \multicolumn{1}{|c}{10.13} & 16.15 & 10.04 & \multicolumn{1}{|c}{14.94} & 18.64 & 14.78 \\
& \multicolumn{1}{|c}{{Renee-I}}  & \multicolumn{1}{|c}{0.59}& 17.78 & 0.65  & \multicolumn{1}{|c}{0.06} & 10.87 & 0.02 & \multicolumn{1}{|c}{0.04} & 14.91 & 0.04 & \multicolumn{1}{|c}{0.05} & 15.81 & 0.03 \\
\bottomrule
\end{tabular}
\end{table*}

\begin{table*}
\centering
\caption{Calibration error metrics on label features based XMLC algorithms. Two consecutive rows, such as NGAME and NGAME-I, represent the measures on the vanilla version of the algorithm and that obtained by post-hoc Isotonic regression respectively.}
\label{tab:LF_table_extended}
\begin{tabular}{
  @{}
  l
  l
  S[table-format=2.2]
  S[table-format=2.2]
  S[table-format=2.2]
  S[table-format=2.2]
  S[table-format=2.2]
  S[table-format=2.2]
  @{}
}
\toprule
 & \textbf{Calibration Measure} &  \multicolumn{1}{|c}{\textbf{ECE-k}} & {\textbf{Brier}} & {\textbf{ACE}} & \multicolumn{1}{|c}{\textbf{ECE-k}} & {\textbf{Brier}} & {\textbf{ACE}} \\
\midrule
 & \multicolumn{1}{c}{\textbf{Dataset}} & \multicolumn{3}{|c}{\textbf{LF-AmazonTitles-131K}} & \multicolumn{3}{|c}{\textbf{LF-WikiSeeAlso-320K}} \\ 
\midrule
\multirow{10}{*}{k=1} & \multicolumn{1}{|c}{{GalaXC}} & \multicolumn{1}{|c}{16.31} & 24.22 & 16.28 & \multicolumn{1}{|c}{10.7} & 18.98 & 11.66 \\
& \multicolumn{1}{|c}{{GalaXC-I}}  & \multicolumn{1}{|c}{0.14} & 20.5 & 0.16 & \multicolumn{1}{|c}{0.21} & 17.19 & 0.22 \\
& \multicolumn{1}{|c}{{NGAME}} & \multicolumn{1}{|c}{11.08} & 20.13 & 11.65  & \multicolumn{1}{|c}{12.73} & 21.1 & 12.87 \\
& \multicolumn{1}{|c}{{NGAME-I}}  & \multicolumn{1}{|c}{0.17} & 17.97 & 0.41 & \multicolumn{1}{|c}{0.16} & 18.9 & 0.26 \\
& \multicolumn{1}{|c}{{Renee}}  & \multicolumn{1}{|c}{24.26} & 26.51 & 23.97 & \multicolumn{1}{|c}{7.77} & 19.43 & 7.77 \\
& \multicolumn{1}{|c}{{Renee-I}}  & \multicolumn{1}{|c}{0.13} & 18.61 & 0.28 & \multicolumn{1}{|c}{0.28} & 18.72 & 0.24\\
& \multicolumn{1}{|c}{{Gandalf}}  & \multicolumn{1}{|c}{16.65} & 20.35 & 16.61 & \multicolumn{1}{|c}{21.45} & 23.99 & 21.47 \\
& \multicolumn{1}{|c}{{Gandalf-I}}  & \multicolumn{1}{|c}{0.16}& 16.76 & 0.19 & \multicolumn{1}{|c}{0.25} & 17.93 & 0.43 \\
\midrule
\multirow{10}{*}{k=3} & \multicolumn{1}{|c}{{GalaXC}} & \multicolumn{1}{|c}{24.09} & 22.99 & 24.07 & \multicolumn{1}{|c}{15.4} & 15.86 & 15.09 \\
& \multicolumn{1}{|c}{{GalaXC-I}}  & \multicolumn{1}{|c}{0.06} & 15.82 & 0.05 & \multicolumn{1}{|c}{0.06} & 12.89 & 0.06 \\
& \multicolumn{1}{|c}{{NGAME}}  & \multicolumn{1}{|c}{16.03} & 17.92 & 15.83 & \multicolumn{1}{|c}{18.65} & 19.33 & 18.39 \\
& \multicolumn{1}{|c}{{NGAME-I}}  & \multicolumn{1}{|c}{0.08} & 14.4 & 0.18 & \multicolumn{1}{|c}{0.1} & 15.21 & 0.12 \\
& \multicolumn{1}{|c}{{Renee}}  & \multicolumn{1}{|c}{17.49} & 19.56 & 16.36 & \multicolumn{1}{|c}{10.11} & 15.79 & 10.11 \\
& \multicolumn{1}{|c}{{Renee-I}}  & \multicolumn{1}{|c}{0.12} & 15.01 & 0.09 & \multicolumn{1}{|c}{0.1} & 14.57 & 0.2\\
& \multicolumn{1}{|c}{{Gandalf}}  & \multicolumn{1}{|c}{13.09} & 16.02 & 12.55 & \multicolumn{1}{|c}{16.03} & 18.26 & 15.09 \\
& \multicolumn{1}{|c}{{Gandalf-I}}  & \multicolumn{1}{|c}{0.07}& 13.74 & 0.08 & \multicolumn{1}{|c}{0.09} & 14.49 & 0.14 \\
\midrule
\multirow{10}{*}{k=5} & \multicolumn{1}{|c}{{GalaXC}} & \multicolumn{1}{|c}{28.78} & 21.77 & 28.77 & \multicolumn{1}{|c}{18.09} & 14.11 & 18.35 \\
& \multicolumn{1}{|c}{{GalaXC-I}}  & \multicolumn{1}{|c}{0.04} & 12.3 & 0.04 & \multicolumn{1}{|c}{0.07} & 10.36 & 0.05 \\
& \multicolumn{1}{|c}{{NGAME}}  & \multicolumn{1}{|c}{19.29} & 15.74 & 18.78 & \multicolumn{1}{|c}{21.3} & 17.52 & 21.29 \\
& \multicolumn{1}{|c}{{NGAME-I}}  & \multicolumn{1}{|c}{0.04} & 11.24 & 0.05 & \multicolumn{1}{|c}{0.09} & 12.43 & 0.04 \\
& \multicolumn{1}{|c}{{Renee}}  & \multicolumn{1}{|c}{13.23} & 14.81 & 13.23 & \multicolumn{1}{|c}{12.29} & 13.14 & 12.29 \\
& \multicolumn{1}{|c}{{Renee-I}}  & \multicolumn{1}{|c}{0.05} & 11.87 & 0.05 & \multicolumn{1}{|c}{0.06} & 11.38 & 0.05\\
& \multicolumn{1}{|c}{{Gandalf}}  & \multicolumn{1}{|c}{10.26}& 12.68 & 9.97 & \multicolumn{1}{|c}{12.85} & 14.58 & 12.78 \\
& \multicolumn{1}{|c}{{Gandalf-I}}  & \multicolumn{1}{|c}{0.05} & 10.97 & 0.06 & \multicolumn{1}{|c}{0.05} & 11.91 & 0.07 \\
\bottomrule
\end{tabular}
\end{table*}